\documentclass[sn-mathphys]{sn-jnl}% Math and Physical Sciences Reference Style
\jyear{2021}%

\usepackage{textcomp,booktabs}
\usepackage{amssymb}% http://ctan.org/pkg/amssymb
\usepackage{pifont}% http://ctan.org/pkg/pifont
\newcommand{\cmark}{\ding{51}}%
\newcommand{\xmark}{\ding{55}}%
\usepackage{xcolor}
\definecolor{SeaGreen4}{RGB}{0,205,102} 
\definecolor{SlateBlue}{RGB}{106,90,205} 
\definecolor{DarkRed}{RGB}{178,34,34}

\theoremstyle{thmstyleone}%
\theoremstyle{thmstyletwo}%

\theoremstyle{thmstylethree}%

\begin{document}

\title[Balanced RGB-Event Video Recognition]{Unleashing the Power of CNN and Transformer for Balanced RGB-Event Video Recognition}

%%=============================================================%%
%% Prefix   -> \pfx{Dr}
%% GivenName    -> \fnm{Joergen W.}
%% Particle -> \spfx{van der} -> surname prefix
%% FamilyName   -> \sur{Ploeg}
%% Suffix   -> \sfx{IV}
%% NatureName   -> \tanm{Poet Laureate} -> Title after name
%% Degrees  -> \dgr{MSc, PhD}
%% \author*[1,2]{\pfx{Dr} \fnm{Joergen W.} \spfx{van der} \sur{Ploeg} \sfx{IV} \tanm{Poet Laureate}
%%                 \dgr{MSc, PhD}}\email{iauthor@gmail.com}
%%=============================================================%%

\author[1]{\fnm{Xiao} \sur{Wang}}
\author[1]{\fnm{Yao} \sur{Rong}}
\author[1]{\fnm{Shiao} \sur{Wang}}
\author[2]{\fnm{Yuan} \sur{Chen}}
\author[3]{\fnm{Zhe} \sur{Wu}}
\author[1]{\fnm{Bo} \sur{Jiang*}}
\author[4]{\fnm{Yonghong} \sur{Tian}}
\author[1]{\fnm{Jin} \sur{Tang}}

\affil[1]{\orgdiv{School of Computer Science and Technology}, \orgname{Anhui University}, \orgaddress{\city{Hefei} \postcode{230601},  \country{China}}}

\affil[2]{\orgdiv{School of Internet}, \orgname{Anhui University}, \orgaddress{\city{Hefei} \postcode{230601},  \country{China}}}

\affil[3]{\orgname{Peng Cheng Laboratory}, \orgaddress{\city{Shenzhen} \postcode{518000},  \country{China}}}

\affil[4]{\orgdiv{School of Computer Science}, \orgname{Peking University}, \orgaddress{\city{Beijing} \postcode{100000},  \country{China}}}
% \thanks{* Corresponding Author: Bo Jiang} 

%%==================================%%
%% sample for unstructured abstract %%
%%==================================%%

\abstract{Pattern recognition based on RGB-Event data is a newly arising research topic and previous works usually learn their features using CNN or Transformer. As we know, CNN captures the local features well and the cascaded self-attention mechanisms are good at extracting the long-range global relations. It is intuitive to combine them for high-performance RGB-Event based video recognition, however, existing works fail to achieve a good balance between the accuracy and model parameters, as shown in Fig.~\ref{firstimage}. In this work, we propose a novel RGB-Event based recognition framework termed TSCFormer, which is a relatively lightweight CNN-Transformer model. Specifically, we mainly adopt the CNN as the backbone network to first encode both RGB and Event data. Meanwhile, we initialize global tokens as the input and fuse them with RGB and Event features using the BridgeFormer module. It captures the global long-range relations well between both modalities and maintains the simplicity of the whole model architecture at the same time. The enhanced features will be projected and fused into the RGB and Event CNN blocks, respectively, in an interactive manner using F2E and F2V modules. Similar operations are conducted for other CNN blocks to achieve adaptive fusion and local-global feature enhancement under different resolutions. Finally, we concatenate these three features and feed them into the classification head for pattern recognition. Extensive experiments on two large-scale RGB-Event benchmark datasets (PokerEvent and HARDVS) fully validated the effectiveness of our proposed TSCFormer. The source code and pre-trained models will be released at \url{https://github.com/Event-AHU/TSCFormer}}

\keywords{Event Camera, RGB-Event based Classification, Temporal Shift, CNN, Transformer }

%%\pacs[JEL Classification]{D8, H51}

%%\pacs[MSC Classification]{35A01, 65L10, 65L12, 65L20, 65L70}

\maketitle

\section{Introduction} \label{sec:intro}
Event-based vision has drawn more and more attention in recent years~\cite{gallego2020eventsurvey,wang2023eventvot,wang2023visevent} due to its significant advantages on high dynamic range, low energy consumption, spatial sparse and temporal dense. It has been introduced in many computer vision tasks, such as object recognition~\cite{li2023semantic, wang2022hardvs}, detection and tracking~\cite{wang2023eventvot, wang2023visevent, gehrig2023recurrentVFormer}, and image enhancement and reconstruction~\cite{teng2022nest, zhu2022eventreconstruction, jiang2023eventLIenhance, ding2023emlb, wu2020probabilistic}. Different from standard RGB sensors which record the light intensity for all pixels in a global exposure way (i.e., synchronously), Event cameras (also called Dynamic Vision Sensors, DVS) emit a spike/event only when the variation of light intensity in a pixel exceeds a given threshold. In other words, the Event cameras asynchronously record the scene using spike/event points similar to point clouds.

\begin{figure*}[!htp]
\center
\includegraphics[width=1\textwidth]{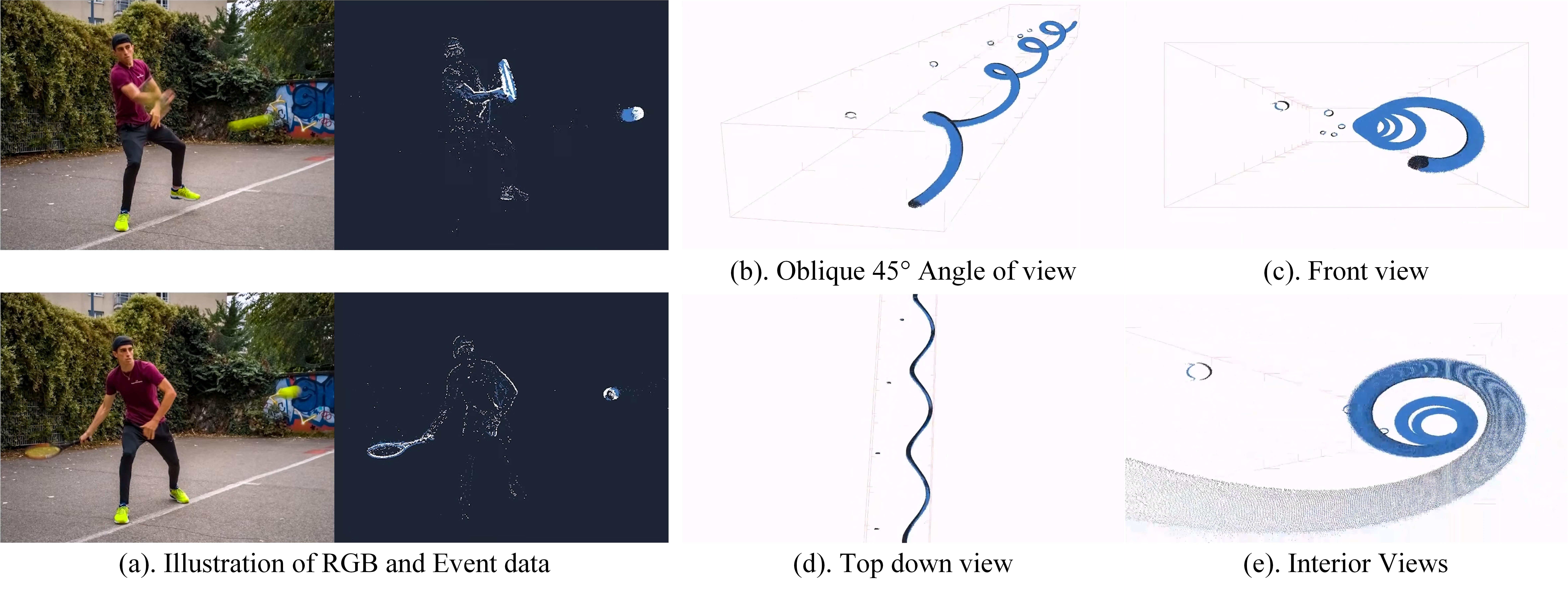}
\caption{Comparison between RGB and Event data (Fig. (a)) and visualizations of Event stream in the 3D views (Fig. (b-e)). 
}      
\label{eventImaging}
\end{figure*}

When the light intensity change at a pixel exceeds the set threshold, the camera records an event. Each event is recorded as a quadruple (x, y, t, p), where x and y represent the spatial coordinates where the event occurs, t is the timestamp, and p represents the polarity of the event. Generally, +1 indicates an increase in brightness, and -1 indicates a decrease in brightness. This characteristic enables the event camera to asynchronously output an event stream related to the change in light intensity and to respond to the dynamic changes in the scene within an extremely short period of time. Event cameras have a high temporal resolution and can focus on the changes in the image on a microsecond time scale. In scenarios where objects are moving at high speed or there are rapid changes in illumination, event cameras are able to capture the changes in objects and record the motion information. Compared with each frame of images recorded by traditional cameras, event cameras only output the pixel information of pixels whose brightness has changed. This enables event cameras to require less bandwidth. A comparison between the RGB frames and Event streams is provided in Fig.~\ref{eventImaging}.

Although the Event camera performs well on the aforementioned aspects, it captures fewer signals when the target object is motionless or static. And only a few event cameras are capable of recording color or detailed texture information, such as the DAVIS346~\cite{Scheerlinck2019CED:}. 
 
These issues make it hard to utilize only Event cameras for some practical scenarios. Hence, some researchers attempt to combine the RGB and Event cameras for a more robust and strong performance~\cite{wang2023sstformer, wang2023visevent, jing2021VSREvent}. In this work, we focus on the task of RGB-Event based pattern recognition which is a fundamental research problem for the Event-based vision. The key challenge is how to encode and fuse these two heterogeneous modalities simultaneously for a better tradeoff between the accuracy and model parameters.

\begin{figure}
\center
\includegraphics[width=0.6\textwidth]{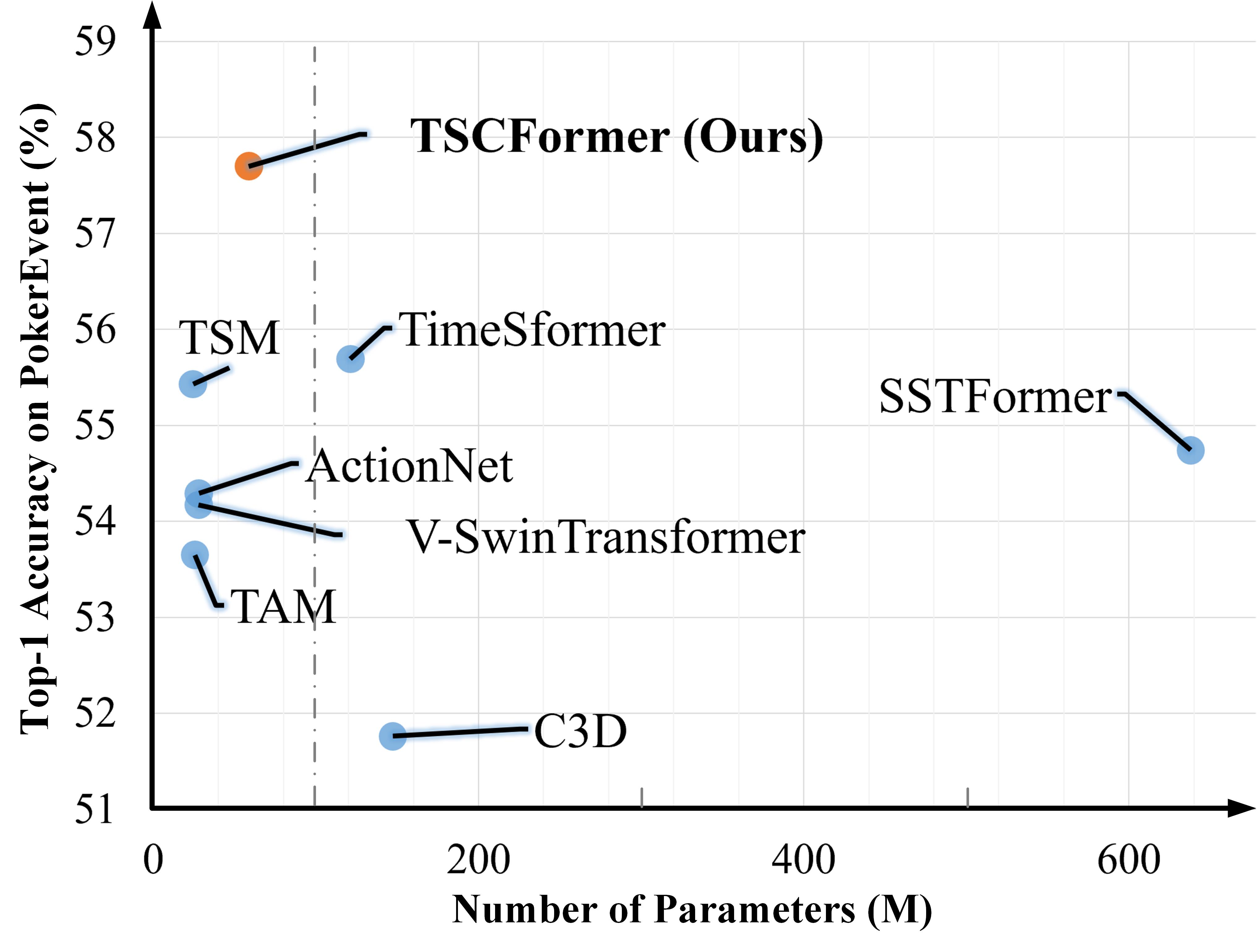}
\caption{A comparison of model parameters and top-1 accuracy between our proposed TSCFormer and other state-of-the-art (SOTA) models on the PokerEvent dataset. It is easy to find that our model achieves a better tradeoff between efficiency and overall performance.}    
\label{firstimage}
\end{figure}

According to our observation, we find that previous RGB-Event based recognition algorithms usually process the dual modalities using CNN (Convolutional Neural Networks), SNN (Spiking Neural Networks), GNN (Graph Neural Networks), or Transformer Networks, due to the diverse representations of Event streams. To be specific, Wang et al.~\cite{wang2023sstformer} propose the SSTFormer which encodes the RGB and Event data using Memory-Support Transformer and SNN and fuse them using a bottleneck mechanism. Although a good performance can be obtained, however, their model contains more than 600M learnable parameters which make it difficult to optimize during actual deployment. For the models with less than 100M parameters like TSM~\cite{lin2019tsm}, TAM~\cite{liu2021tam}, and Video SwinTransformer~\cite{liu2022video}, as noted in Fig.~\ref{firstimage}, we can find that their results are relatively poor on the public PokerEvent~\cite{wang2023sstformer} benchmark dataset. These observations inspired us to think about \textit{how to achieve a good balance between the model parameters and recognition performance for the RGB-Event based recognition?}

In this paper, we propose a novel framework that aggregates the temporal shift CNN and lightweight BridgeFormer modules for RGB-Event based pattern recognition, termed TSCFormer.   As shown in Fig.~\ref{framework}, we transform the event streams into event images and adopt stem networks to embed both RGB and Event data. Then, temporal-shift-based CNN modules are proposed to encode the RGB and Event feature maps to get the local feature representations. A set of global tokens is randomly initialized and fused with local CNN features using the BridgeFormer layer. The output features will be concatenated with RGB and Event CNN representations after feature transformation via a fully connected layer and reshape operation. Similar local-global feature aggregation procedures are also conducted for other temporal shift CNN blocks. Finally, we flatten the local CNN features of the two modalities and concatenate them with global tokens, then, map them into category labels using the classifier for recognition. According to Fig.~\ref{firstimage}, we can find that our proposed TSCFormer achieves a better balance between the model parameters and recognition performance.

To sum up, the key contributions of this paper can be summarized as the following three aspects: 

$\bullet$ We propose a novel RGB-Event based recognition framework, termed TSCFormer, which achieves a better tradeoff between the model parameters and recognition performance. It encodes the RGB/Event data using temporal shift CNN blocks and boosts the interactions of multimodal features with global BridgeFormer layers. 

$\bullet$ We propose novel F2V/F2E modules that boost the information propagation between local convolutional RGB/Event features and global Transformer representations. 

$\bullet$ Extensive experiments are conducted on two large-scale RGB-Event based recognition benchmark datasets, specifically, we set new state-of-the-art performances on these datasets, 53.04\% and 57.70\% on HARDVS and PokerEvent, respectively.

The rest of this paper is organized as follows: 
In section~\ref{sec:relatedworks}, we give a brief introduction to the most related works. Then, we will introduce our proposed TSCFormer framework in section~\ref{sec:Method} with a focus on the overview, input representation, temporal shift CNN sub-network, BridgeFormer sub-network, classification head, and loss function. We conduct extensive experiments on two datasets which will be introduced in section~\ref{datasetsMetric}. The implementation details, comparisons with SOTA models, ablation studies, parameter analysis, visualizations, and limitation analysis are all described in this part. Finally, we conclude this paper and propose possible research directions for the next steps in section~\ref{conclusions}.

\section{Related Works}  \label{sec:relatedworks} 

In this section, we give a brief review of Event-based Recognition, RGB-Event based Recognition, and Transformer Networks. More related works can be found in the following surveys~\cite{gallego2020eventsurvey, han2022transformersurvey, wang2023MMPTMs} and paper list~\footnote{\url{https://github.com/Event-AHU/Event_Camera_in_Top_Conference}}.

\subsection{Event-based Recognition} 
In recent years, Event-based technology has garnered widespread attention in the fields of computer vision and machine learning. Significant progress has been made in Event-based recognition, providing valuable experience and theoretical foundations for the development of Event-based recognition technology. 
To be specific, Cho et al.~\cite{cho2023label} propose a joint learning framework for Event-based object recognition and image reconstruction, which conducts joint learning without requiring paired images and labels. 
Innocenti et al.~\cite{innocenti2021temporal} propose a concise representation of event data called Temporal Binary Representation, which leverages the conversion of binary event sequences into frames to encode both spatial and temporal information. Their approach enables the adjustment of information loss and memory footprint, rendering it suitable for real-time applications.
The authors of~\cite{Kim_2022_CVPR} propose a test time adaptation algorithm for event-based object recognition, referred to as Ev-TTA. It enables the classifier to adapt to new, unseen, and widely varying environments during the testing phase. 
Cannici et al.~\cite{cannici2019attention} propose an attention mechanism algorithm that monitors event activity in a scene and extracts patches from reconstructed frames, for image classification, capable of recognizing objects within the reconstructed frames. 
Wang et al.~\cite{Wang_2019_CVPR} propose an Event-based gait recognition method, Ev-Gait. This algorithm effectively eliminates noise from the event stream by enhancing motion consistency and utilizes deep neural networks to recognize gait from asynchronous and sparse event data. 
Liu et al.~\cite{liu2021event} propose a hierarchical SNN architecture for Event-based action recognition using motion information. 
Compared with these works, our proposed TSCFormer is a multi-modal framework that models the RGB and Event data simultaneously using temporal shift based CNN and BridgeFormer module.

\subsection{RGB-Event based Recognition} 
Multi-modal based research has been widely exploited in many downstream tasks~\cite{islam2022maven, mai2022multimodalbottleneck, wang2022mfgnet}. By integrating RGB images and Event data, RGB-Event based recognition can provide a more robust recognition performance. In comparison to traditional Event-based recognition, RGB-Event recognition places an emphasis on the fusion of visual information and Event data to achieve a comprehensive and enriched understanding of scenes and actions. Previous work provides an empirical and theoretical basis for the development of Event-based recognition techniques. 
Specifically, 
Huang et al.~\cite{huang2022vefnet} propose to utilize a cross-attention module on the RGB-Event bimodal data to endow the network with cross-modal feature selection capabilities, while the self-attention module combines nearby local features. 
Wang et al.~\cite{wang2023sstformer} propose an RGB frame-event recognition framework for the simultaneous fusion of RGB frames and event streams, termed SSTFormer. It utilizes a memory-supported Transformer network for RGB frame encoding and a spiking neural network for raw event stream encoding. 
Li et al.~\cite{li2023semantic} introduce a Semantic-Aware Frame-Event fusion based pattern recognition framework, called SAFE, based on a large-scale vision-language pre-trained model. This model integrates RGB frames, Event data, and semantic label sets into a unified framework. 
%%%% 
Compared to previous works, the TSCFormer proposed in this work processes the RGB and Event data using temporal shift CNN and further enhances the local CNN features using the BridgeFormer module. Note that, the BridgeFormer proposed for the global feature learning is lightweight and achieves a better tradeoff between the accuracy and model complexity.

\subsection{Transformer Network}  
In recent years, the Transformer~\cite{vaswani2017attention} network is increasingly applied to fuse information from different modalities~\cite{Truong_2021_ICCV, radford2021learning, He_2022_CVPR, Neimark_2021_ICCV}, effectively processing and integrating diverse types of data. This demonstrates the potential of the Transformer architecture in handling multi-modal data. As a result, there has been growing interest in exploring and advancing multi-modal Transformer networks for a wide range of applications. 
%%%% 
Peng et al.~\cite{peng2021conformer} propose a hybrid neural network that combines the CNN and Transformer for RGB-based representation learning, termed ConvFormer. It encodes the local and global features by inputting the RGB image into the CNN and Transformer, respectively. 
Bottleneck fusion based multi-modal framework is proposed in \cite{nagrani2021AttBottleNeck} by Arsha et al. which achieves a better balance between the performance and computational cost. 
Zhang et al.~\cite{zhang2023cmx} explore RGB-X semantic segmentation for the first time using five multi-modal sensor data combinations: RGB-Depth, RGB-Thermal, RGB-Polarization, RGB-Event, and RGB-LiDAR. The authors propose an RGB-X semantic segmentation framework called CMX, which features cross-modal feature calibration and fusion modules, interleaved cross-attention, and mixed-channel embedding to enhance global reasoning. 
Zhao et al.~\cite{zhao2022tftn} design a Transformer-based Fusion Module (TFM) to explore the relationship between the spectral representations of hyper-spectral and RGB data within this framework.
The author of~\cite{bozic2021transformerfusion} employs the Transformer network to learn the fusion of time-domain multi-view features to obtain a hierarchical structure in a coarse to fine manner. 
Wang et al.~\cite{wang2021transformer} introduce a transformer-based RGB-D saliency detection framework, which globally integrates cross-modal and scale features simultaneously. 
Zhou et al.~\cite{zhou2022multispectral} propose a multi-spectral fusion transformer network to highlight the intra-spectral correlation and inter-spectral complementary of RGB-T. 
Li et al.~\cite{li2021trear} propose a feature fusion block that pays mutual attention to learning a joint feature representation for classification. 
The authors of~\cite{Wang_2022_CVPR} propose a multi-modal token fusion method (TokenFusion) that effectively integrates multiple modalities, allowing the transformer to learn correlations. 
Prakash et al.~\cite{Prakash_2021_CVPR} propose a Multi-Modal Fusion Transformer (TransFuser) to integrate the global context of the 3D scene into the feature extraction layers of various modalities. 
%%%% 
Inspired by these works, in this paper, we propose a novel framework that facilitates the information propagation between local convolutional RGB/Event features and global Transformer representations in a relatively lightweight way.

\begin{figure*} 
\center
\includegraphics[width=1\textwidth]{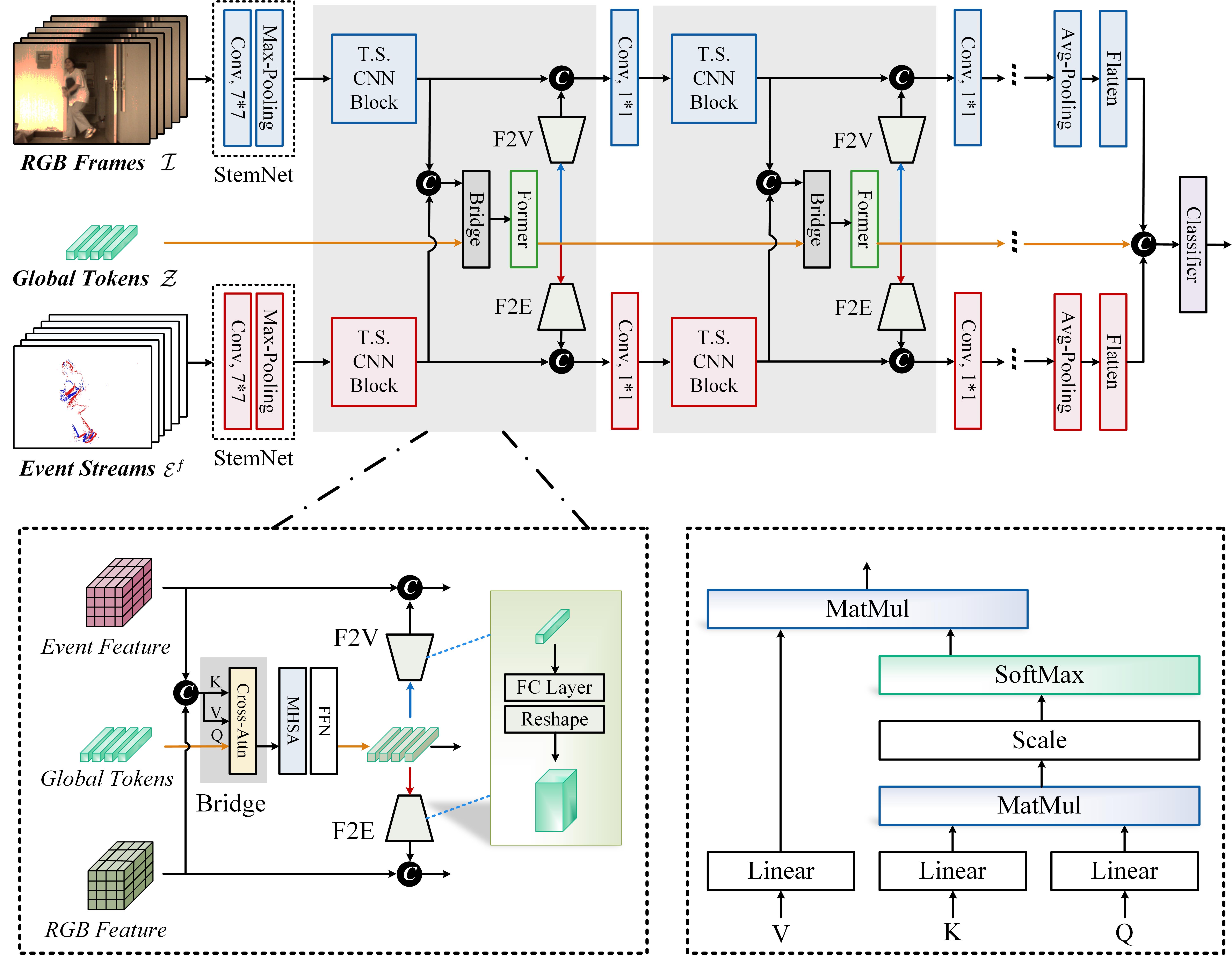}
\caption{
\textbf{An overview of our proposed RGB-Event based pattern recognition framework, termed TSCFormer.} Specifically, we first adopt a stem network to process the RGB frames and event streams into feature embeddings. Then, the local features of RGB and Event data can be obtained using temporal-shift-based CNN (ResNet50~\cite{He_2016_CVPR} is used in this work). To get the global representations, we randomly initialize a set of global tokens and feed them into the BridgeFormer module. Therefore, the output global tokens will be fed into RGB and Event branch for fusion, and the subsequent BridgeFormer module for further information aggregation. Finally, we fuse all three branches and feed them into a fully connected layer for classification. 
}  
\label{framework}
\end{figure*}

\section{Our Proposed Method}  \label{sec:Method} 

In this section, we will first present the overview of our proposed RGB-Event recognition framework TSCFormer, and then introduce the input representations of RGB frames and Event streams. And we will further dive into the details of our framework and primarily focus on the Temporal Shift CNN Sub-Network and the BridgeFormer Sub-Network. Afterward, we will discuss the classification head and the loss function used during the training phase.

\subsection{Overview} 
\begin{figure} 
\center
\includegraphics[width=0.8\textwidth]{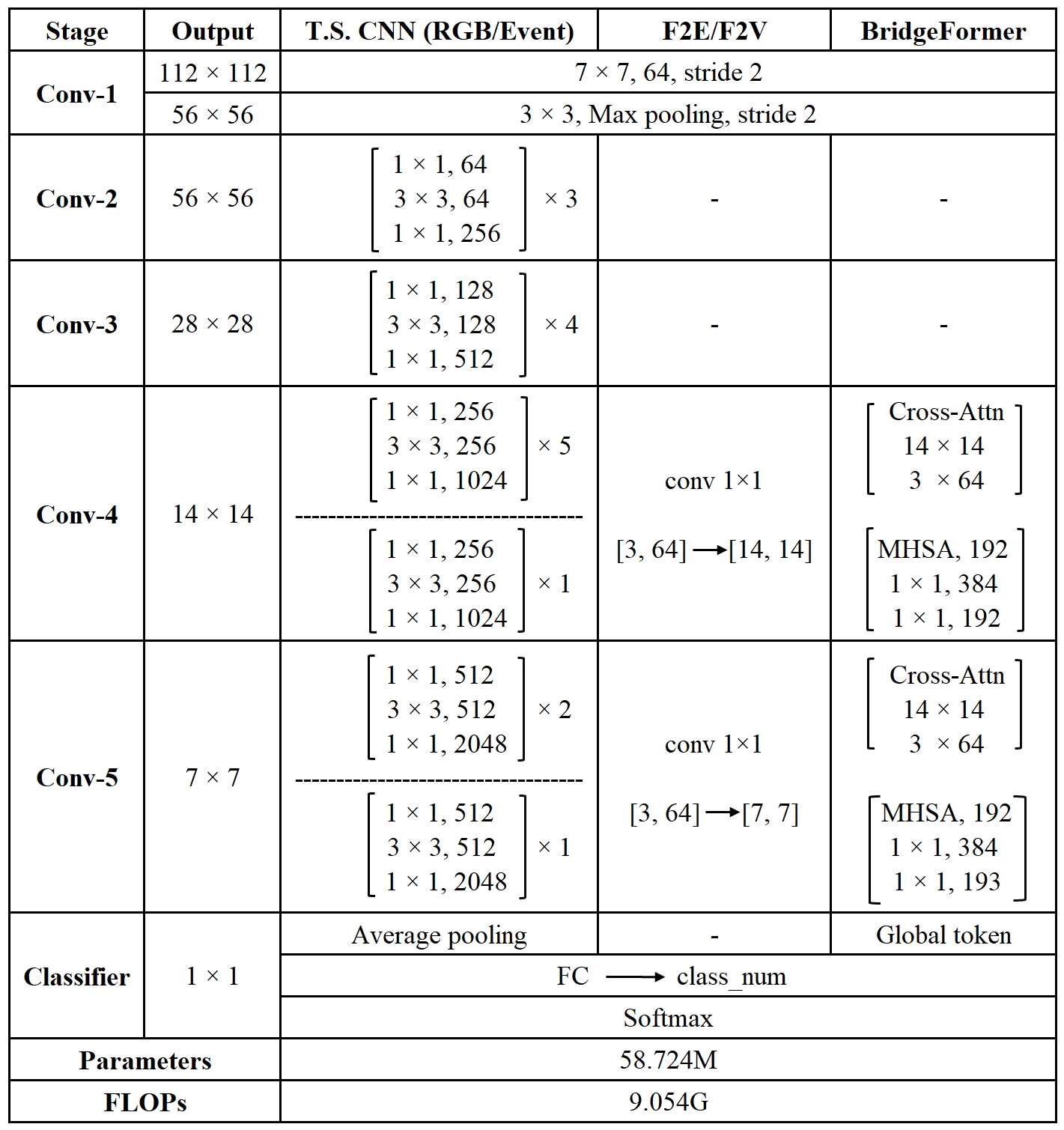}
\caption{The detailed network architecture of our proposed TSCFormer framework.}      
\label{networkarchitecture}
\end{figure}

As shown in Fig.~\ref{framework}, given the RGB frames and Event streams, we first adopt a stem net which contains a convolutional layer and max-pooling layer to obtain the feature embeddings. Then, the two feature maps are fed into temporal-shift-based CNN modules to get the local feature representations. Meanwhile, a set of global tokens is randomly initialized and fed into the BridgeFormer layer. Note that, the local RGB/Event feature maps are concatenated and transformed into key and value tokens, and the global tokens are transformed into query tokens. Then, a cross-attention layer is adopted to fuse the local and global features and a Transformer layer is used to learn the global tokens. The output features will be concatenated with RGB and Event CNN representations after feature transformation via a fully connected layer and reshape operation. After that, we adopt a convolutional layer with kernel size $1 \times 1$ to fuse these features and feed them into the subsequent temporal shift CNN blocks for a similar operation. Finally, we flatten the features of the two modalities and map them into category labels using the classifier. Fig.~\ref{networkarchitecture} shows the specific details of the network architecture. More detailed computing procedures will be introduced in the following subsections.

\subsection{Input Representation} 
Given the RGB frames $\mathcal{I} \in \mathbb{R}^{B \times T \times C \times H \times W} = \{ I_1, I_2, ..., I_T \}$ and Event streams $\mathcal{E}^{p} = \{ e_{1}, e_{2}, ..., e_{M} \}$, $T$ and $M$ denotes the number of video frames and event points, each point $e$ is a quadruple $\{x, y, t, p\}$, where $(x, y)$ denotes the spatial coordinates, $t$ and $p$ denotes the time stamp and polarity. Here, $B$ denotes the batch size, $\{C, H, W\}$ denotes the dimension of channel, height, and width, respectively. 
We first transform the event streams into event images $\mathcal{E}^{f} = \{ E_1, E_2, ..., E_T \}$ based on the timestamp of RGB frames \footnote{\url{https://github.com/wangxiao5791509/VisEvent_SOT_Benchmark/blob/main/scripts/read_aedat4.py}}. Then, we feed the RGB and Event frames into temporal-shift CNN branch for the local feature extraction. We adopt two stem networks (a convolutional layer with filter $7 \times 7$ and max-pooling layer) to embed the RGB and Event frames into their feature representations $F_I \in \mathbb{R}^{B \times T \times C \times H \times W}$ and $F_E \in \mathbb{R}^{B \times T \times C \times H \times W}$, respectively. Meanwhile, we also randomly initialize a set of global tokens $\mathcal{Z} \in \mathbb{R}^{B \times L \times D}$ as the input of the BridgeFormer module for local-global feature aggregation.

\begin{figure} 
\center
\includegraphics[width=0.8\textwidth]{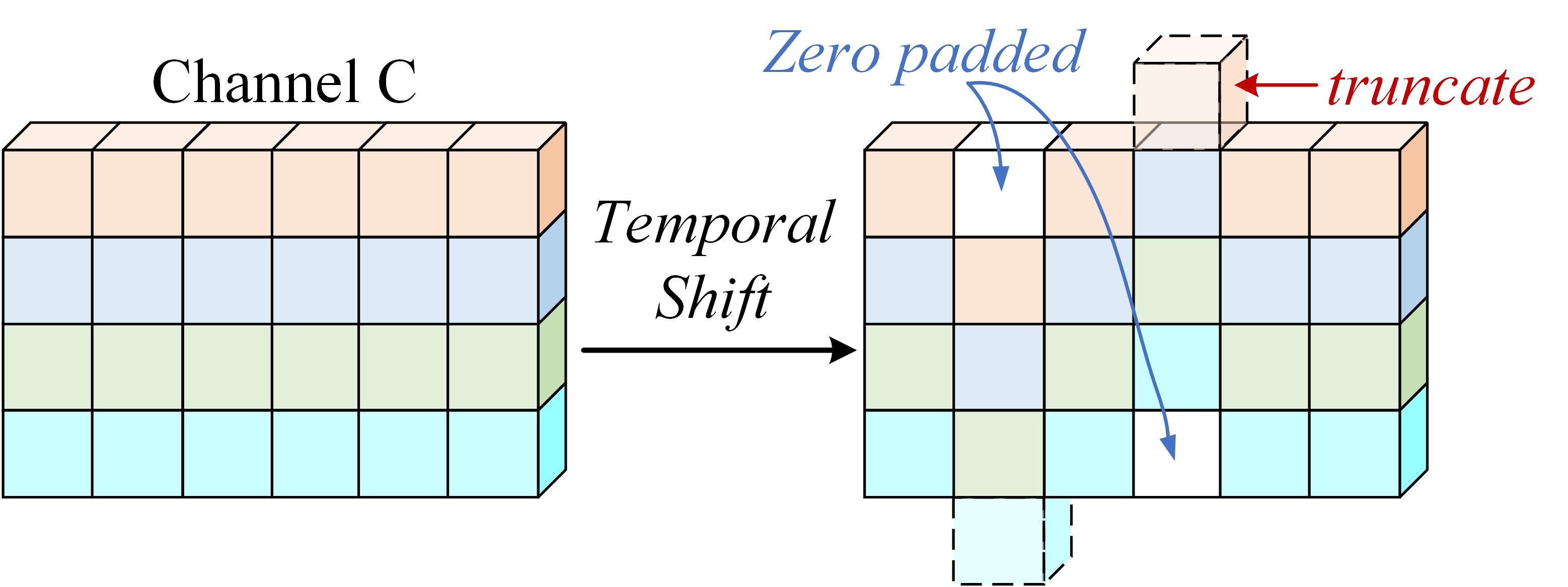}
\caption{Illustration of Temporal Shift (T.S.) Operator.}     
\label{TSM}
\end{figure}

\subsection{Temporal Shift CNN Sub-Network}  
After we get the embedded feature maps of RGB and Event frames, i.e., $F_I$ and $F_E$, we feed them into temporal-shift-based CNN modules, respectively. As noted in TSM~\cite{lin2019tsm}, this simple operation can achieve the performance of 3D CNN but maintain 2D CNN’s complexity. Specifically speaking, the temporal-shift module will shift part of the channels along the temporal dimension, therefore, facilitating information exchange between neighboring frames. An illustration of the temporal shift module is provided in Fig.~\ref{TSM}. 
%%%%
Then, the shifted features $\hat{F}_I$ and $\hat{F}_E$ will be fed into the CNN blocks (ResNet50~\cite{He_2016_CVPR} is adopted in this work) for enhanced feature extraction. In our practical implementation, four residual blocks are adopted for the feature extraction. For each block $i \in \{1, 2, 3, 4\}$, we concatenate the RGB and Event features [$\hat{F}_I^i$, $\hat{F}_E^i$] along the channel dimension and feed into the BridgeFormer module which will be introduced in the next subsection.

\subsection{BridgeFormer Sub-Network} 
The aforementioned temporal-shift CNN block learns the local features well, however, the global long-range relations can be better obtained using the self-attention-based Transformer network as validated in many previous works~\cite{peng2021conformer}-~\cite{Prakash_2021_CVPR}. In this work, we propose a novel BridgeFormer that efficiently fuses and enhances the RGB and Event features. More specifically, we randomly initialize a set of global tokens $\mathcal{Z} \in \mathbb{R}^{B \times L \times D} = \{ Z_1, Z_2, ..., Z_B \}$ and treat it as the query feature $\mathcal{Q}$. The concatenated RGB-Event features [$\hat{F}_I^i$, $\hat{F}_E^i$] are transformed into key and value features, i.e., $\mathcal{K}$ and $\mathcal{V}$. Then, a cross-attention layer is utilized to fuse these tokens and the detailed procedure can be formulated as: 
\begin{equation}
    \label{crossattention}  
    CrossAtten(\mathcal{Q}, \mathcal{K}, \mathcal{V}) = Softmax(\frac{\mathcal{Q} \mathcal{K}^T}{\sqrt{d}}) \mathcal{V},  
\end{equation} 
where $d$ is the dimension of token representation. Considering the self-attention's powerful ability to model long-term distance relationships, we process the output tokens into $Q$, $K$, and $V$, and feed into the Multi-Head Self-Attention (MHSA) layer whose key operation is the self-attention layer, i.e., 
\begin{equation}
    \label{selfattention} 
    SelfAtten(Q, K, V) = Softmax(\frac{{Q} {K}^T}{\sqrt{d}}) {V}. 
\end{equation} 
We also introduce the Feed-Forward Network (FFN) to get the enhanced feature tokens inspired by ViT~\cite{dosovitskiy2020image} and Transformer~\cite{vaswani2017attention}. The output will be fused with RGB and Event CNN features, respectively, and also passed into the next BridgeFormer module for further processing.

When aggregating the CNN local features and global tokens, we introduce the F2V and F2E modules for the feature transformation from BridgeFormer to RGB branch and Event branch, respectively. As illustrated in Fig.~\ref{framework}, this feature transformation module contains a fully connected layer and a reshape operator. The reshaped features will be processed using a convolution layer with kernel size $1 \times 1$ to keep the same dimension with feature maps transferred between multiple vanilla CNN blocks. Therefore, we can utilize the pre-trained weights on the ImageNet dataset to initialize our CNN blocks. It is worth noting that when fusing the global tokens and RGB/Event CNN features, we tried the cross-attention scheme. However, a sub-optimal performance is obtained compared with the simple concatenate operation.

\subsection{Classification Head \& Loss Function} 
After the fourth CNN block, we adopt the average pooling layer to get the representative feature map of RGB and Event data. Then, we flatten the feature maps into feature vectors and concatenate them with global tokens for classification. The classifier we used contains one fully connected layer.

As the RGB-Event based pattern recognition is a multi-category classification problem, in this work, we adopt the widely used cross-entropy loss function to calculate the distance between the predicted label distribution $\hat{y}$ and ground truth $y$, i.e., 
\begin{equation}
    \label{lossFunction} 
    \mathcal{L} = -\left[y\log\hat{y}+ (1-y)\log(1-\hat{y})\right] . 
\end{equation}

\section{Experiments}  \label{sec:Experiments} 
In this section, we will first introduce the datasets and evaluation metrics used in our experiments. Then, the comprehensive implementation details are reported to help other researchers to reproduce this work. After that, we report our recognition results and compare them with other state-of-the-art models on two large-scale benchmark datasets. We also conduct extensive experiments to validate the effectiveness of each module in our framework and find a better setting and parameters for final recognition. We also give a visualization of the features learned by our model and discuss the limitations of this work.

\subsection{Datasets and Evaluation Metric} \label{datasetsMetric}

In our experiments, two large-scale frame-event based pattern recognition datasets are used, including {HARDVS}~\cite{wang2022hardvs} and {PokerEvent}~\cite{wang2023sstformer}. Specifically speaking, the HARDVS~\footnote{\url{https://github.com/Event-AHU/HARDVS}} dataset is collected using a DVS346 event camera, which concentrates on recognizing human activities such as walking, running, and crouching. It consists of 300 classes and includes 96,908 RGB-Event samples. It is divided into training and testing subsets, with 64,522 and 32,386 samples, respectively. The PokerEvent~\footnote{\url{https://github.com/Event-AHU/SSTFormer}} dataset focuses on recognizing character patterns in poker cards. It consists of 114 classes and includes 24,415 RGB-Event samples recorded using a DVS346 event camera. The dataset is divided into training and testing subsets, with 16,216 and 8,199 samples, respectively. The \textbf{top-1 accuracy} and \textbf{top-5 accuracy} are adopted to evaluate the performance of our model and other state-of-the-art recognition algorithms.

\subsection{Implementation Details} 
In our experiments, we set the batch size to 8 and train our framework for 30 epochs on both the HARDVS and PokerEvent datasets. We use the SGD~\cite{Bottou2012Stochastic} optimizer with an initial learning rate of 0.001 for the training process. Our parameter scheduler is \textit{MultiStepLR}, and when the training epoch reaches 25, the learning rate decreases by a factor of 0.1. During the training phase, we first perform \textit{MultiScaleCrop} on the input images of two modalities, with scales set to 1, 0.875, 0.75, and 0.66. Then, we resize them to $224 \times 224$, and finally apply a flip with ratio of 0.5. We use RGB and Event videos for the two branches in the framework, with 8 frames selected for each modality, totaling 16 frames. We apply a temporal shift operation to the input feature, with the number of divisions for the shift set to 8. We select the widely used cross-entropy loss as the loss function to quantify the disparity between the ground truth and our model's predictions. This loss function is commonly employed in classification tasks and is effective in penalizing the deviations of predicted probabilities from the actual class labels. By utilizing the cross-entropy loss, we aimed to optimize parameters of our model to minimize the discrepancy between the predicted and actual outcomes, thereby enhancing the predictive performance. Our code is implemented using PyTorch~\cite{paszke2019pytorch} based on Python. All the experiments were conducted on a server with RTX3090 GPUs.

\subsection{Comparison on Public Benchmarks} 
To objectively evaluate the performance of our proposed method, we conduct extensive experiments on HARDVS and PokerEvent datasets and compare our results with other state-of-the-art recognition models in Table \ref{HARDVS_acc} and Table \ref{PokerEvent_acc}.

\noindent 
\textbf{Results on HARDVS Dataset~\cite{wang2022hardvs}.~} 
As shown in Table~\ref{HARDVS_acc}, we compare our proposed TSCFormer with 6 SOTA models, including 3D CNN based model C3D~\cite{tran2015learning}, ResNet based model (TSM~\cite{lin2019tsm},  X3D~\cite{feichtenhofer2020x3d} and ESTF~\cite{wang2022hardvs}), Transformer-based methods (TimeSformer~\cite{bertasius2021space} and SSTFormer~\cite{wang2023sstformer}). 
It is easy to find that our TSCFormer achieves the best recognition performance on this dataset, i.e., $53.04\%$ on the top-1 evaluation metric. For the top-5 metric, our model also achieves the best accuracy, i.e., $62.67\%$, which exceeds the second one (TSM~\cite{lin2019tsm}) by a margin $+0.55\%$. These results fully validated the effectiveness of our proposed local-global feature aggregation mechanism for RGB-Event pattern recognition.                                        

\begin{figure} 
\center
\includegraphics[width=3.3in]{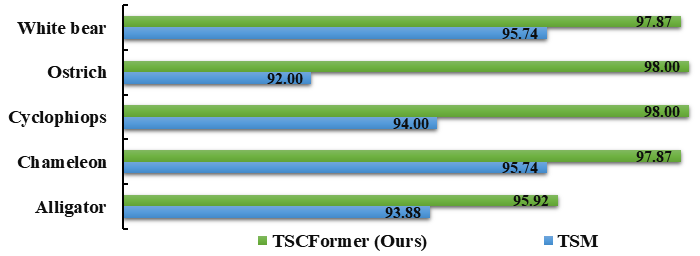}
\caption{The improvement of Top1 accuracy using TSCFormer compared to TSM on five classes of PokerEvent dataset.}  
\label{improvement}
\end{figure}

\noindent 
\textbf{Results on PokerEvent Dataset~\cite{wang2023sstformer}.~} 
As shown in Table~\ref{PokerEvent_acc}, we compare our model with 9 SOTA algorithms on the PokerEvent dataset. 
We can find that our model achieves $57.7\%$ on the top-1 evaluation metric which surpasses the latest recognition methods V-SwinTrans~\cite{liu2022video} and MVIT~\cite{li2022mvitv2}. Also, our model beats the baseline TSM~\cite{lin2019tsm} which obtained $55.43\%$ on this dataset. The detailed comparisons of recognition results of five classes (\textit{White bear, Ostrich, Cyclophiops, Chameleon, Alligator}) on PokerEvent dataset are illustrated in Fig.~\ref{improvement}. We can find that our model significantly improves the recognition results over the TSM model on these 5 categories. To be specific, the top-1 accuracy are improved up to $97.87\%$, $98.00\%$, $98.00\%$, $97.87\%$, $95.92\%$, respectively. These experimental results further proved the effectiveness of our proposed local-global feature fusion and enhancement modules for RGB-Event based video recognition.

\begin{table}
\center
\small   
\caption{Results on the HARDVS dataset (Event+RGB). 
The best and second results are highlighted in \textcolor{red}{\textbf{red}} and \textcolor{blue}{\textbf{blue}}.} 
\label{HARDVS_acc}
\resizebox{\columnwidth}{!}{ 
\begin{tabular}{l|l|c|c|c|c|c}
\hline 
\textbf{No.} & \textbf{Algorithm} & \textbf{Publish}  & \textbf{Backbone}  & \textbf{Modality} &\textbf{Top-1}  &\textbf{Top-5} \\
\hline
\#01 &\textbf{C3D}~\cite{tran2015learning}     &ICCV-2015   &3D-CNN & Event+RGB &50.88 &56.51    \\ 
\hline
\#02 &\textbf{TSM}~\cite{lin2019tsm}     &ICCV-2019   &ResNet-50 & Event+RGB &52.58 &\textcolor{blue}{\textbf{62.12}}    \\ 
\hline
\#03 &\textbf{TimeSformer}~\cite{bertasius2021space}    &ICML-2021   &ViT & Event+RGB &51.57 &58.48    \\ 
\hline
\#04 &\textbf{X3D}~\cite{feichtenhofer2020x3d}     &CVPR-2020   &ResNet & Event+RGB &47.38 &51.42    \\ 
\hline
\#05 &\textbf{ESTF}~\cite{wang2022hardvs}     &AAAI-2024   &ResNet18 & Event+RGB &49.93  &55.77    \\ 
\hline
\#06 &\textbf{SSTFormer}~\cite{wang2023sstformer}     &arXiv-2023   &SNN-Former & Event+RGB &\textcolor{blue}{\textbf{52.97}} &60.17    \\ 
\hline
\#07 &\textbf{TSCFormer (Ours)}     &-   &TSCFormer & Event+RGB &\textcolor{red}{\textbf{53.04}} &\textcolor{red}{\textbf{62.67}}    \\ 
\hline
\end{tabular}
}
\end{table}

\begin{table}
\center
\small   
\caption{Results on the PokerEvent dataset (Event+RGB). 
The best and second results are highlighted in \textcolor{red}{\textbf{red}} and \textcolor{blue}{\textbf{blue}}.} 
\label{PokerEvent_acc}
\resizebox{\columnwidth}{!}{ 
\begin{tabular}{l|l|c|c|c|c}
\hline 
\textbf{No.} & \textbf{Algorithm}   & \textbf{Publish}  & \textbf{Backbone} &\textbf{Modality} &\textbf{Top-1}  \\
\hline
\#01 &\textbf{C3D}~\cite{tran2015learning}     &ICCV-2015   &3D-CNN &Event+RGB &51.76    \\ 
\hline
\#02 &\textbf{TSM}~\cite{lin2019tsm}     &ICCV-2019   &ResNet-50 &Event+RGB &55.43    \\ 
\hline
\#03 &\textbf{ACTION-Net}~\cite{wang2021action}     &CVPR-2021   &ResNet-50 &Event+RGB &54.29 \\ 
\hline
\#04 &\textbf{TAM}~\cite{liu2021tam}     &ICCV-2021   &ResNet-50 &Event+RGB &53.65    \\ 
\hline
\#05 &\textbf{V-SwinTrans}~\cite{liu2022video}  &CVPR-2022   &Swin Transformer &Event+RGB &54.17 \\ 
\hline
\#06 &\textbf{TimeSformer}~\cite{bertasius2021space}     &ICML-2021   &ViT &Event+RGB &\textcolor{blue}{\textbf{55.69}}    \\ 
\hline
\#07 &\textbf{X3D}~\cite{feichtenhofer2020x3d}     &CVPR-2020   &ResNet &Event+RGB &51.75    \\ 
\hline
\#08 &\textbf{MVIT}~\cite{li2022mvitv2}     &CVPR-2022   &ViT   &Event+RGB &55.02    \\ 
\hline
\#09 &\textbf{SSTFormer}~\cite{wang2023sstformer}     &arXiv-2023   &SNN-Former &Event+RGB &54.74    \\ 
\hline
\#10 &\textbf{TSCFormer (Ours)}      &-   &TSCFormer  &Event+RGB &\textcolor{red}{\textbf{57.70}}    \\ 
\hline
\end{tabular}
}
\end{table}

\subsection{Ablation Study} 
To better validate the superiority of our proposed method, we present ablation experiments in this section. This includes analyzing the benefits of each component in our method and assessing the impact of different inputs on the final results. 

\noindent \textbf{Component Analysis.~} 
The key components of our proposed TSCFormer framework are the Temporal Shift CNN Sub-Network (T.S.), Cross-attention based Bridge Module, and Self-attention based Transformer layer. As shown in Table~\ref{PokerEvent_Component}, we conduct the component analysis on the PokerEvent dataset by removing each module separately. Specifically, we can find that our model achieves $57.70\%$ on the top-1 accuracy when all these modules are used. The results are dropped to $56.71\%$, $56.06\%$, and $56.49\%$ when removing the temporal shift module, Transformer, and Bridge module, respectively. These results demonstrate that all the modules contribute to our final recognition performance. 
%%%% 
Also, we find that when only T.S. is used, the result is $56.88\%$, however, the result is dropped when introducing the Transformer layers only. If we use the Transformer and Bridge module simultaneously, the results can be improved to $57.70\%$, which demonstrates that the cross-attention based Bridge module is a vital component in the TSCFormer. When only the CNN is used for recognition, the top-1 accuracy is $55.93\%$ which validated the effectiveness of the temporal shift module in our framework. We analyze the impact on the classification performance of removing the F2V and F2E modules. As shown in Table~\ref{F2V_F2E}, when the F2V and F2E modules are used, the top-1 accuracy of the model is 57.70. However, when the F2V and F2E modules are removed, the top-1 accuracy of the model drops to 56.82. The experimental results demonstrate the effectiveness of the proposed F2V and F2E modules.

\noindent \textbf{Compare of Only Event Inputs.~} 
Table~\ref{only_event} shows the experimental results of these video recognition approaches when only event data is utilized. Our model achieves better performance compared to other video recognition approaches. The experimental results demonstrate the effectiveness of the proposed method.

\noindent \textbf{Influence of Single- and Multi-modal Inputs.~} 
To validate the effect of different modalities and the combination of RGB and Event data on recognition performance, we conduct experiments using various input combinations. As shown in Table \ref{different inputs}, using only event data achieved the lowest accuracy, indirectly indicating the difficulty of extracting features from Event data compared to RGB data. Additionally, using only RGB data achieved an accuracy of $56.64\%$, while combining both modalities yielded the highest result. This suggests the beneficial impact of multi-modal data on improving recognition performance and the effectiveness of our proposed method in thoroughly extracting and integrating features from two modalities.

\begin{table}
\center
\small   
\caption{Component Analysis on the PokerEvent dataset. } 
\label{PokerEvent_Component}
\begin{tabular}{c|cccc|c}
\hline 
\textbf{No.}  & \textbf{CNN} & \textbf{T.S.}  & \textbf{Former}  & \textbf{Bridge}  & \textbf{Top-1} \\
\hline
\textbf{01}   &\textcolor{SeaGreen4}{\cmark}   &\textcolor{SeaGreen4}{\cmark}   &\textcolor{SeaGreen4}{\cmark}     &\textcolor{SeaGreen4}{\cmark}     &57.70  \\
\textbf{02}   &\textcolor{SeaGreen4}{\cmark}   &\textcolor{DarkRed}{\xmark}   &\textcolor{SeaGreen4}{\cmark}     &\textcolor{SeaGreen4}{\cmark}     &56.71  \\
\textbf{03}   &\textcolor{SeaGreen4}{\cmark}   &\textcolor{SeaGreen4}{\cmark}   &\textcolor{DarkRed}{\xmark}         &\textcolor{SeaGreen4}{\cmark}         &56.06   \\
\textbf{04}   &\textcolor{SeaGreen4}{\cmark}   &\textcolor{SeaGreen4}{\cmark}   &\textcolor{SeaGreen4}{\cmark}     &\textcolor{DarkRed}{\xmark}         &56.49  \\
\hline 
\textbf{05}   &\textcolor{SeaGreen4}{\cmark}   &\textcolor{SeaGreen4}{\cmark}   &\textcolor{DarkRed}{\xmark}         &\textcolor{DarkRed}{\xmark}         &56.88   \\ 
\textbf{06}   &\textcolor{SeaGreen4}{\cmark}   &\textcolor{DarkRed}{\xmark}   &\textcolor{DarkRed}{\xmark}     &\textcolor{DarkRed}{\xmark}     &55.93  \\

\hline
\end{tabular}
\end{table}

\begin{table}
\center
\small   
\caption{Results of the model with and without the F2V and F2E modules.} 
\label{F2V_F2E}
\begin{tabular}{l|c|c}
\hline 
\textbf{No.}   &\textbf{Method}  &\textbf{Top-1}  \\
\hline
\textbf{01}       &\ with F2V and F2E     &\ 57.70    \\ 
\textbf{02}  &\ without F2V and F2E     &\ 56.82    \\
\hline
\end{tabular}
\end{table}

\begin{table}
\center
\small   
\caption{Results of utilize only event data on the HARDVS dataset.}
\label{only_event}
\begin{tabular}{l|c|c|c}
\hline 
\textbf{No.}   &\textbf{Algorithm} &\textbf{Modality}  &\textbf{Top-1}  \\
\hline
\textbf{01}  &\ C3D     &\ Event  &\ 50.52   \\ 
\textbf{02}  &\ TSM     &\ Event  &\ 52.63   \\
\textbf{03}  &\ TimeSformer     &\ Event  &\ 50.77   \\
\textbf{04}  &\ X3D     &\ Event  &\ 45.82   \\
\textbf{05}  &\ ESTF     &\ Event  &\ 51.22   \\
\textbf{06}  &\ SSTFormer     &\ Event  &\ 51.87   \\
\textbf{07}  &\textbf{TSCFormer(Ours) }     &\ Event   &\textbf{52.70}    \\
\hline
\end{tabular}
\end{table}

\begin{table}
\center
\small   
\caption{Results of different inputs on PokerEvent dataset. } 
\label{different inputs}
\begin{tabular}{l|c|c}
\hline 
\textbf{Input}   &\textbf{Backbone}  &\textbf{Top-1}  \\
\hline
\textbf{(RGB, RGB)}       &\ TSCFormer     &\ 56.64    \\ 
\textbf{(Event, Event)}  &\ TSCFormer     &\ 54.73    \\
\textbf{(RGB, Event)}  &\ TSCFormer     &\ 57.70    \\
\hline
\end{tabular}
\end{table}

\begin{figure*}[!htp]
\center
\includegraphics[width=1\textwidth]{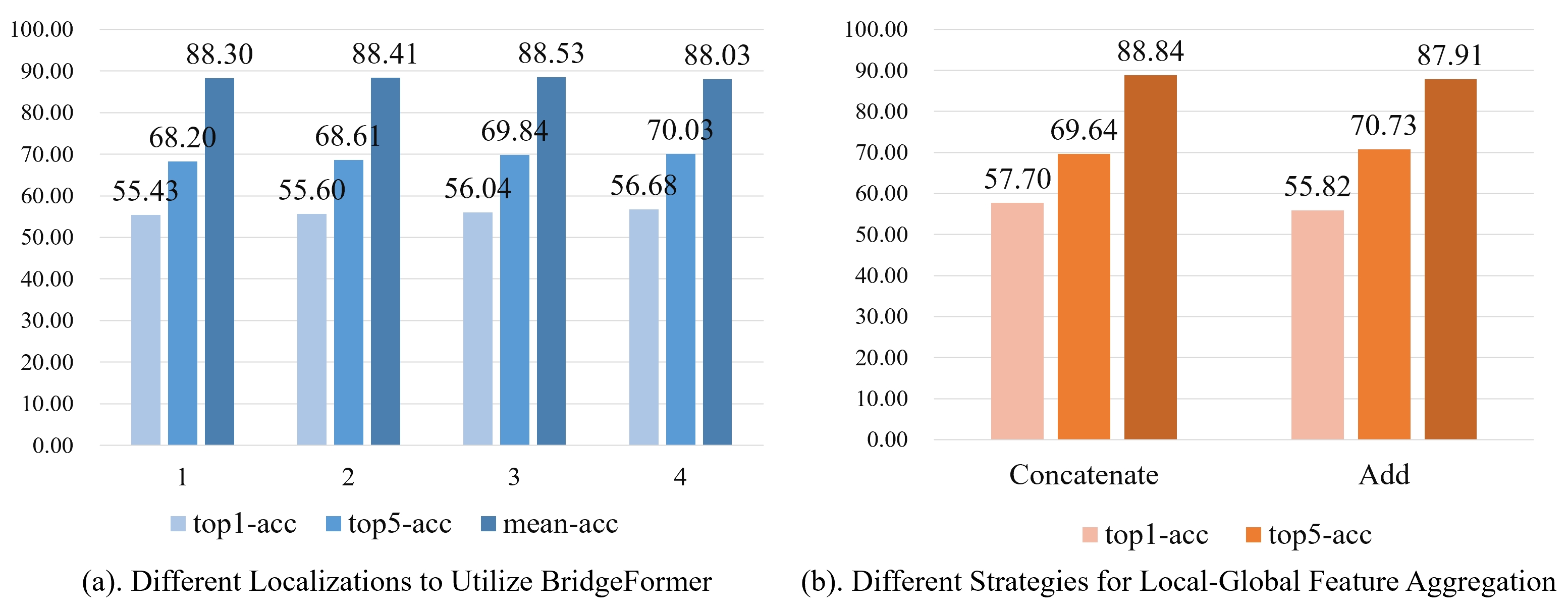} 
\caption{Results of different locations for injecting BridgeFormer module and different fusion strategies.}   
\label{other_analysis}
\end{figure*}

\noindent \textbf{Influence of Different Localizations to Utilize BridgeFormer.~} 
As shown in Fig.~\ref{other_analysis} (a), we attempt to insert the BridgeFormer module into different layers of our TSCFormer framework, i.e., the $1^{th}, 2^{th}, 3^{th}$, and $4^{th}$ T.S. CNN block. It is easy to find that the best top-1 accuracy ($56.68\%$) can be obtained when the $4^{th}$ block is selected on the PokerEvent dataset. The mean accuracy will be at its peak when the $3^{th}$ block is used and the result is $88.53\%$. These results demonstrate that the aggregation of local and global features will be better in the late stages of our framework. We believe this maybe caused by the fact that the high-level semantic features will be more suitable for global long-range relations mining.

\noindent \textbf{Concatenate or Add when Fusing Local-Global Features?~}  
When fusing the local CNN features and the global Transformer tokens, different fusion strategies can be adopted. To be specific, we tried the concatenate and add operators and found that a better result can be obtained when the concatenate operation is used. As shown in Fig.~\ref{other_analysis} (b), the mean accuracy of concatenate and add operations is $88.84\%$ and $87.91\%$, respectively, on the PokerEvent dataset. Therefore, we select the concatenate operation when designing our TSCFormer framework.

\begin{figure*}[!htp]
\center
\includegraphics[width=1\textwidth]{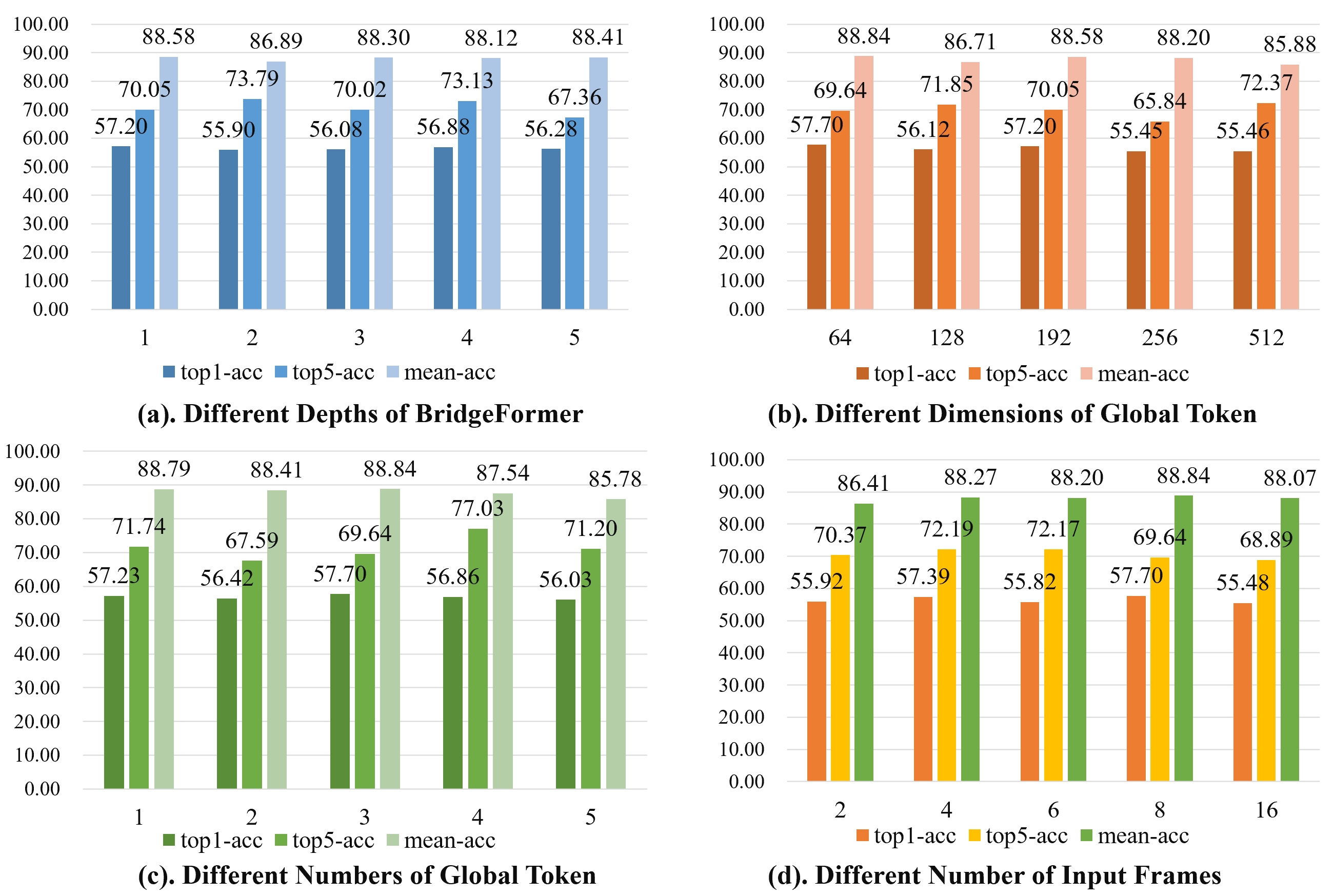} 
\caption{Parameter Analysis on the Global Tokens used in BridgeFormer module.}   
\label{paraAnalysis1}
\end{figure*}

\subsection{Parameter Analysis} 
In this part, we discuss different settings of key parameters in our framework and report the detailed results in Fig.~\ref{paraAnalysis1}.

\noindent \textbf{Analysis on the Depth of BridgeFormer Modules.~} 
To better obtain global long-range relations, a Transformer structure is employed in our method. It is commonly believed that the depth of a neural network has an impact on its performance, and generally, performance improves with the increase in depth. Therefore, we trained our BridgeFormer with different depths, including 1, 2, 3, 4, and 5. The top-1, top-5, and average accuracy are reported in Fig.~\ref{paraAnalysis1} (a). Surprisingly, deeper depth does not necessarily lead to better performance. In terms of average accuracy, the highest accuracy is achieved when the depth is 1. Therefore, in this paper, we set the depth to 1.

\noindent \textbf{Analysis on Various Dimensions of Global Tokens.~}
In this paper, we introduce several global tokens $\mathcal{Z}$ as a query feature, which is a crucial part of cross-attention computation. Hence, we observe its impact on model performance by setting different token dimensions. As shown in Fig.~\ref{paraAnalysis1} (b), we set the dimensions to 64, 128, 192, 256, and 512, and find that the top-5 accuracy can be slightly improved with the increase of dimension but the average accuracy reaches its highest value of 88.84 when the dimension is 64.

\noindent \textbf{Analysis on Different Number of Global Tokens.~}
Additionally, we also evaluate the effect of different numbers of global tokens on recognition performance, as depicted in Fig.~\ref{paraAnalysis1} (c). With the increase of the token quantity, top-1 accuracy varies slightly, reaching $77.03\%$ when the number is 4. The average accuracy, however, reaches the highest value of $88.84\%$ at number 3. This demonstrates that changes in token number and dimension have a certain impact on model performance, and it is necessary to set appropriate values to optimize the model's performance.

\noindent \textbf{Analysis on Different Number of Input Frames.~} 
In our RGB-Event recognition framework, the input includes RGB frames and event streams, hence, we discuss whether the number of input frames affects the recognition performance. As shown in Fig.~\ref{paraAnalysis1} (d), inputting 2, 4, 6, 8, and 16 frames, it can be observed that initially, with an increase in the number of input frames, accuracy improves, but then it decreases. The highest average accuracy is achieved when the number of frames is 8. This indicates that different frames can provide different information, and as the number of frames increases, the model can extract more features. However, an excessive number of frames may lead to information redundancy, resulting in adverse effects.

\begin{figure} 
\center
\includegraphics[width=1\textwidth]{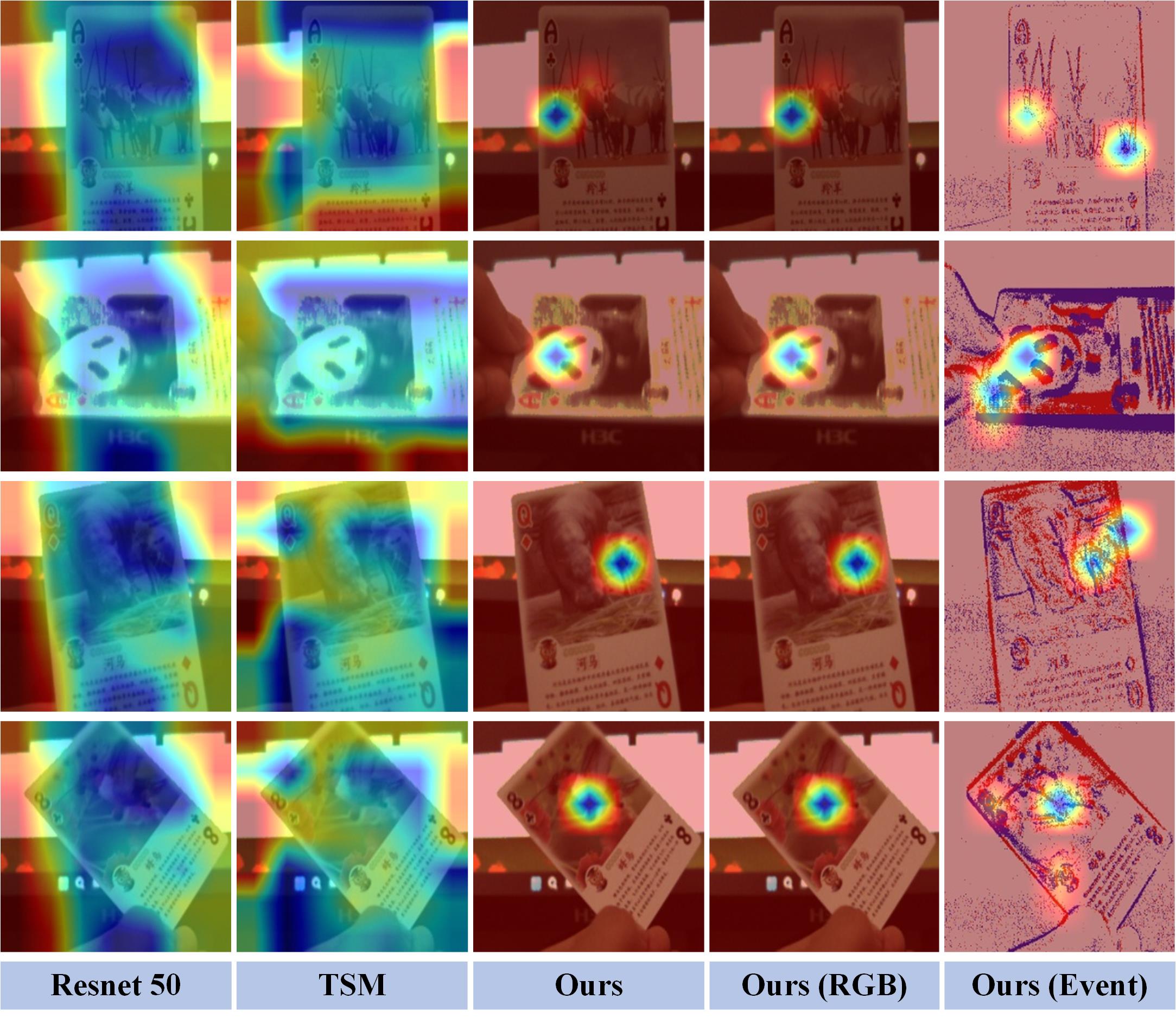}
\caption{Visualization of feature maps obtained using ResNet50, TSM, and our proposed TSCFormer. }  
\label{visual_featMAPs1}
\end{figure}

\begin{figure*} 
\center
\includegraphics[width=1\textwidth]{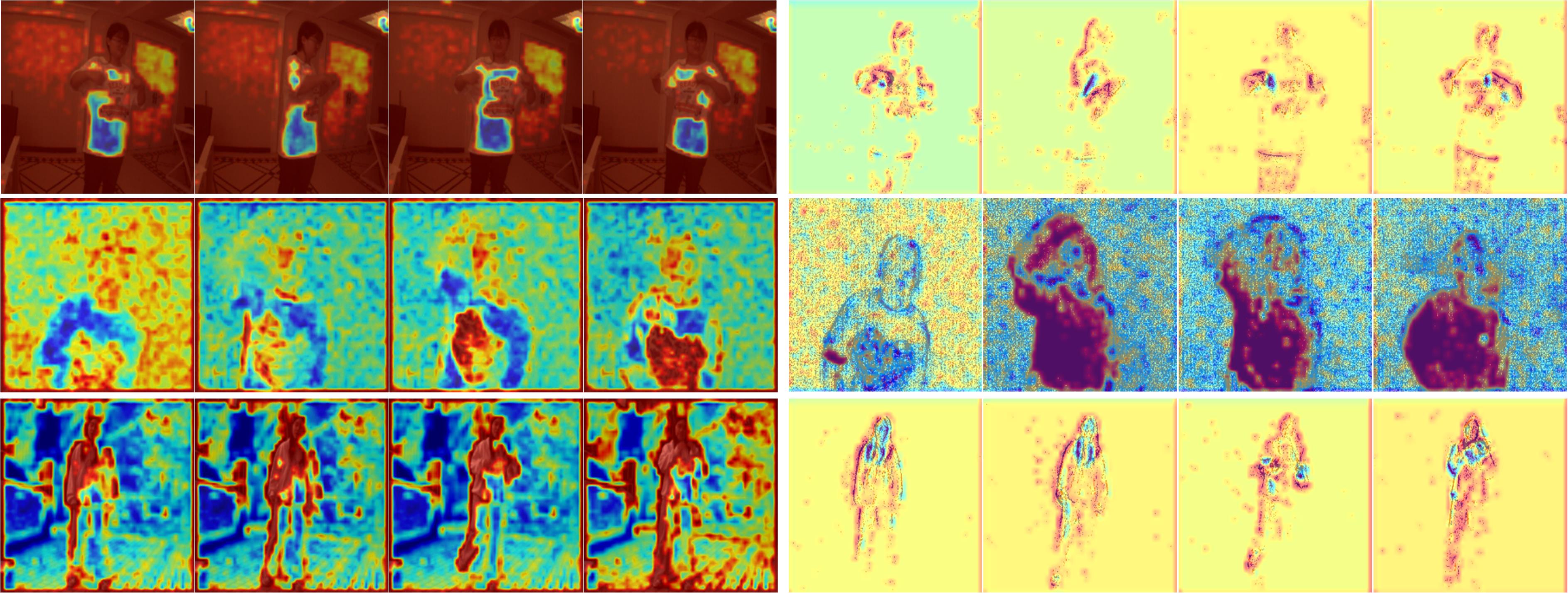}
\caption{Visualization of feature maps obtained using our TSCFormer on samples from HARDVS dataset. }  
\label{visual_featMAPs2}
\end{figure*} 

\begin{figure*} 
\center
\includegraphics[width=1\textwidth]{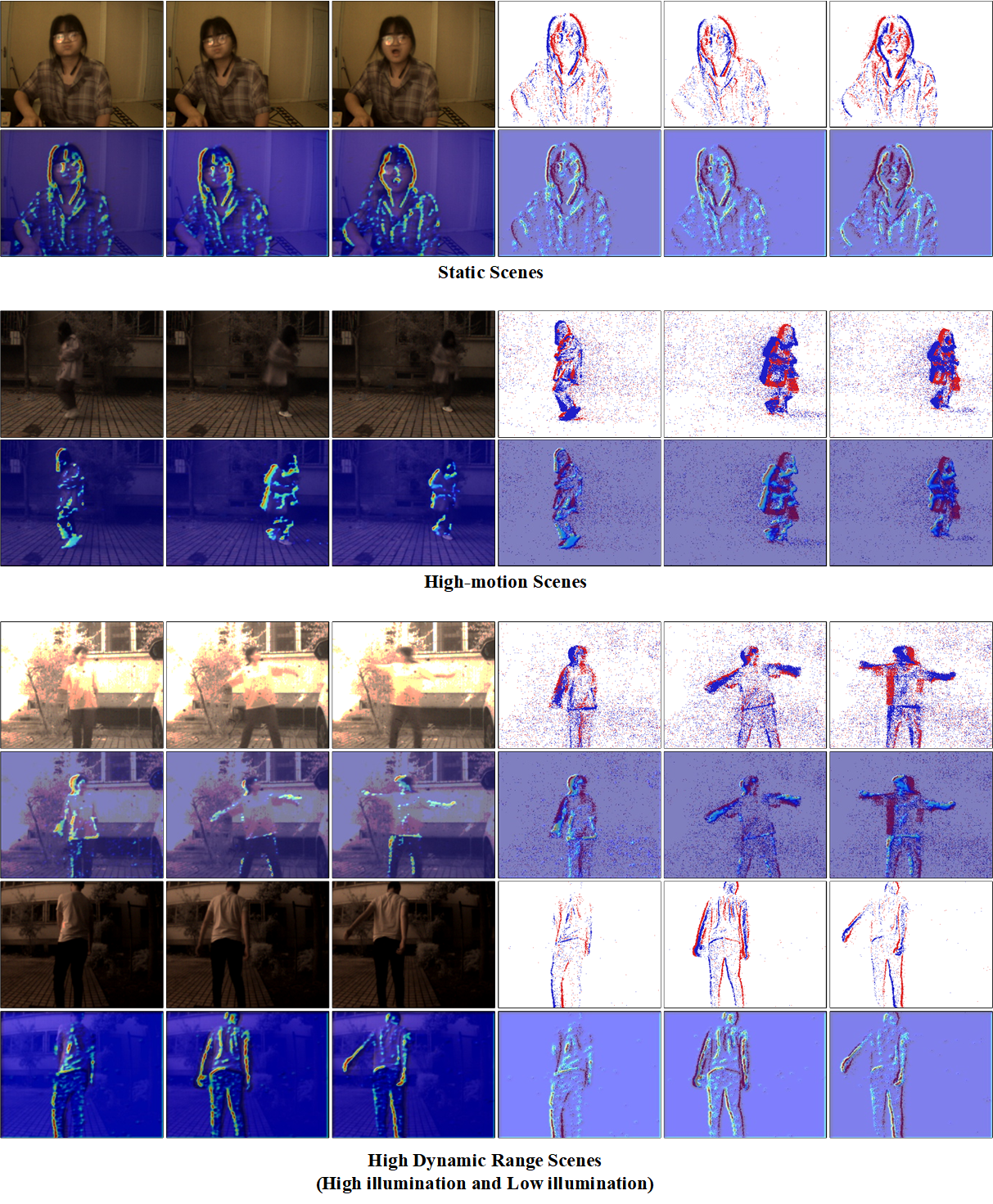}
\caption{The visual comparison examples of static scenes, high-motion scenarios and high dynamic range environments. }
\label{visual_featMAPs_sences}
\end{figure*}

\begin{figure*} 
\center
\includegraphics[width=1\textwidth]{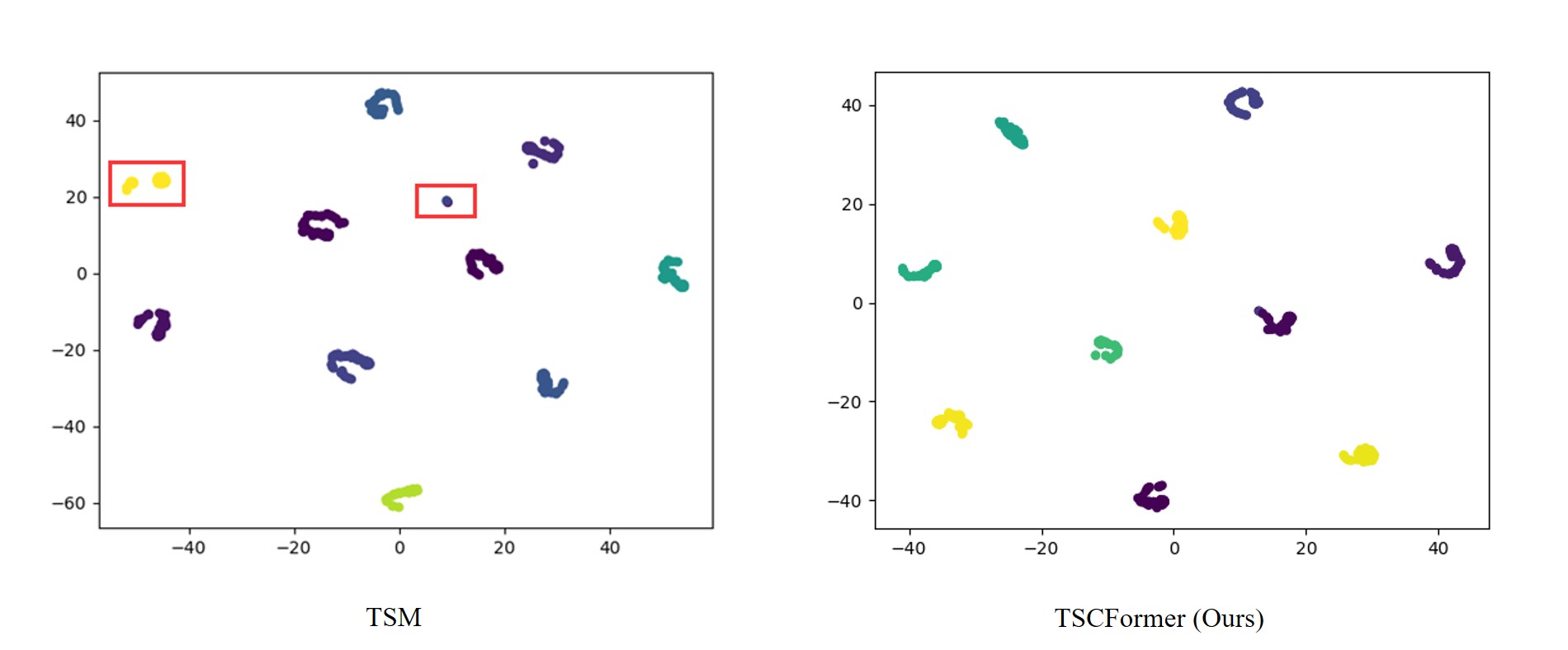}
\caption{Visualization of feature distribution of TSM and TSCFormer (Ours) on PokerEvent. }  
\label{visual_aggregation}
\end{figure*}

\begin{figure*} 
\center
\includegraphics[width=1\textwidth]{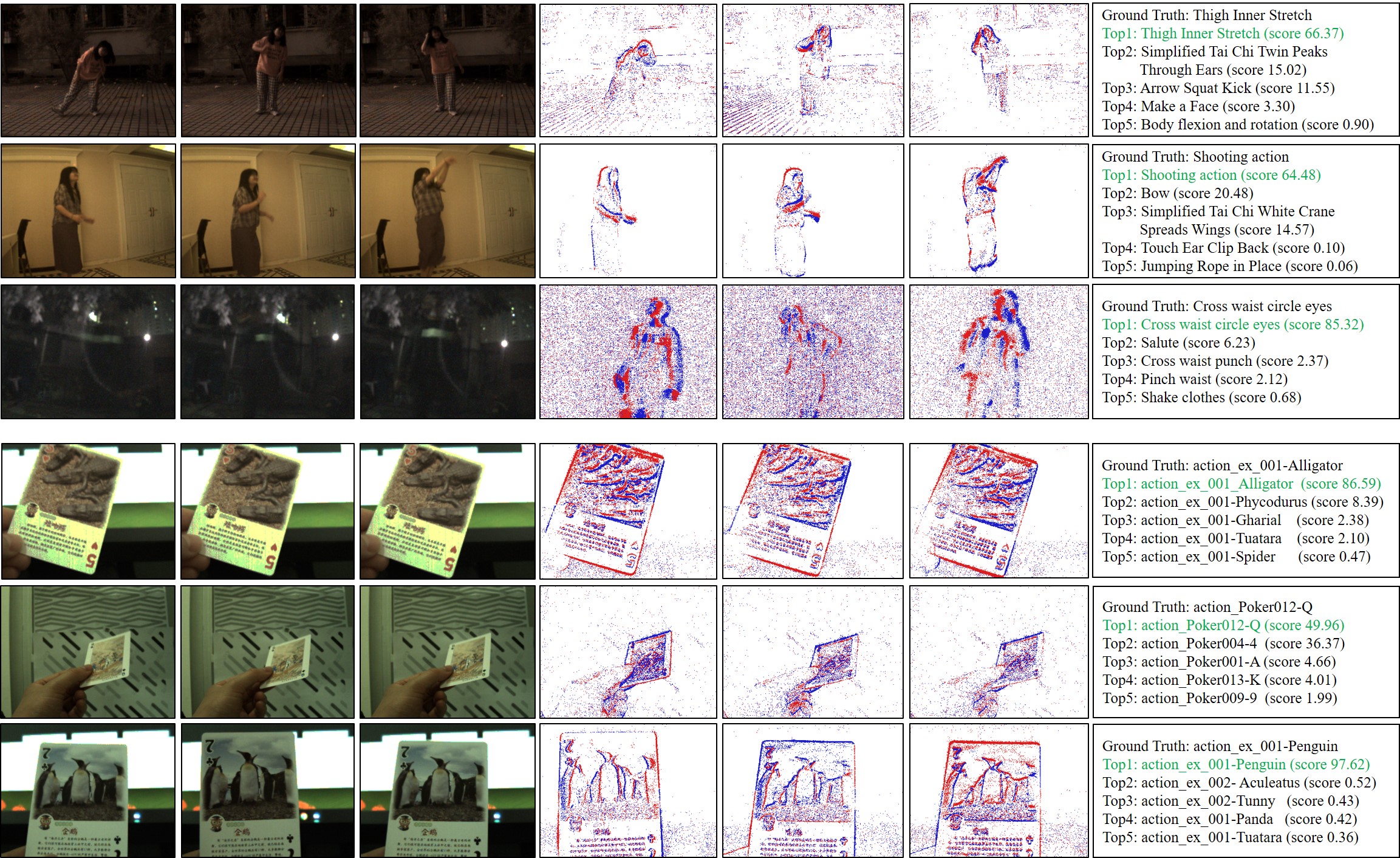}
\caption{Visualization of the top-5 predicted labels using our model on the HARDVS and PokerEvent datasets.}  
\label{top5_vis}
\end{figure*}

\begin{figure*} 
\center
\includegraphics[width=1\textwidth]{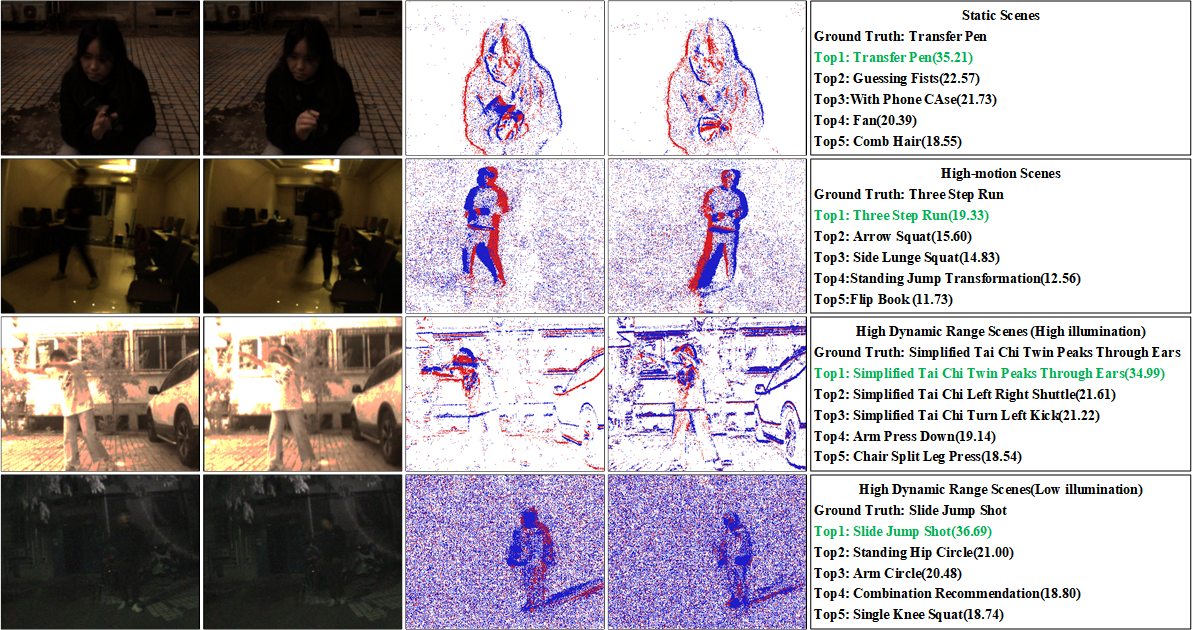}
\caption{The top-5 prediction examples of static scenes, high-motion scenarios and high dynamic range environments.}  
\label{top5_sences}
\end{figure*}

\subsection{Visualization}

In addition to the aforementioned quantitative analysis, in this part, we also give some visualizations to help the readers better understand our TSCFormer framework. Specifically speaking, the visualization of feature maps, feature distributions, and top-5 predictions are all provided in the subsequent paragraphs. 

\noindent 
\textbf{Feature Maps and Distribution.~} 
As shown in Fig.~\ref{visual_featMAPs1}, we give a comparison between the feature maps of ResNet50, TSM, and our TSCFormer. We can find that our model achieves more focused attention on the Poker samples. For the feature maps obtained on the HARDVS dataset, as shown in Fig.~\ref{visual_featMAPs2}, we can find that our model also can locate the key regions that will help recognize human activities. 

As shown in the Fig.~\ref{visual_featMAPs_sences}, the first and second rows represent the RGB and event samples and their corresponding visualization effects under static scenes. The third and fourth rows represent the situations under high-motion scenarios. The fifth to eighth rows display the cases in a high dynamic range environment. Among them, the fifth and sixth rows indicate that the samples are under high illumination, and the seventh and eighth rows indicate that the samples are under low illumination. In multiple challenging scenarios of Fig.~\ref{visual_featMAPs_sences}, the sample visualization effects focus on the people in the scenes and their behaviors and actions, indicating that it is effective to perform video classification by combining RGB and event data. 

%%%%% 
As illustrated in Fig.~\ref{visual_aggregation}, we visualize the feature distribution of TSM and our TSCFormer on the PokerEvent dataset. We showcase the distances between different classes (randomly 10 classes are selected). It is evident that the feature aggregation effect of our TSCFormer is more precise than that of TSM, effectively minimizing the distribution error observed in the baseline. This further underscores the superiority of our method over the baseline approach.

\noindent 
\textbf{Top-5 Predictions.~} 
As shown in Fig.~\ref{top5_vis}, we provide six groups of samples and corresponding predictions on both the HARDVS and PokerEvent datasets. From the samples in the first 3 rows, we can find that the human activities can be accurately identified even in the low-light scenarios. Due to the effectiveness of event data, we can achieve favorable results on the HARDVS dataset. The later 3 rows display the performance of our model on the PokerEvent dataset. We can find that our model accurately identifies the true categories of Pokers, aligning closely with the ground truth. 

As shown in Fig.~\ref{top5_sences}, the samples in the first row represent the situation under static scenes. The model predicts the top five classification categories, and the top-1 prediction result is consistent with the ground truth. The samples in the second row are under high-motion scenarios. The samples in the third and fourth rows are in a high dynamic range environment, representing the environments with high illumination and low illumination respectively. In multiple scenarios, the top five prediction results of the model are consistent with the ground truth, which proves that it is beneficial to perform video classification by combining RGB and event data.

\subsection{Limitation Analysis} 
Although our proposed TSCFormer achieves better results on both HARDVS and PokerEvent datasets, however, our framework can still be improved from the following aspects: 
Firstly, our TSCFormer contains the ResNet as the backbone network and is initialized using the weights pre-trained on the ImageNet dataset. However, it failed to reap the benefits of pre-trained large models, and a lot of work has demonstrated its strong generalization ability~\cite{radford2021learning, wang2023MMPTMs}. 
Secondly, we formulate the RGB-Event pattern recognition problem as a mapping from vision modality to semantic labels in this work, however, there still exists the semantic gaps between these modalities. For example, the BLIP-2~\cite{li2023blip2} demonstrates that the introduction of Q-Former will bridge such semantic gaps significantly. 
%%%% 
In our future works, we will consider further improving our framework from the aforementioned two aspects.

\section{Conclusion and Future Works} \label{conclusions}
In this paper, we propose a novel RGB-Event based pattern recognition framework, termed TSCFormer, which is a relatively lightweight CNN-Transformer network architecture. The CNN (ResNet50) is adopted to encode the RGB and Event data and the temporal shift operation is also proposed to boost the interaction between nearby frames. To make our model lightweight enough, we propose to capture the global long-range relations using BridgeFormer module which takes the random initialized global tokens as the input. Then, the enhanced features will be projected and fused into the RGB and Event CNN blocks,  respectively in an interactive manner using F2E and F2V modules. We conduct similar operations for other CNN blocks to achieve adaptive fusion and local-global feature enhancement under different resolutions. These features are concatenated and fed into a classifier for final recognition. We conducted extensive experiments on two large-scale RGB-Event datasets, i.e., PokerEvent and HARDVS, which fully validated the effectiveness of our proposed TSCFormer.  
%%%%% 
In our future works, we will consider further improving our framework from the perspective of large-scale pre-training and strategies for bridging the semantic gaps between multiple modalities.

\backmatter

% \bmhead{Supplementary information}

% If your article has accompanying supplementary file/s please state so here.

% Authors reporting data from electrophoretic gels and blots should supply the full unprocessed scans for key as part of their Supplementary information. This may be requested by the editorial team/s if it is missing.

% Please refer to Journal-level guidance for any specific requirements.

\section*{Acknowledgments}
This work is supported by the National Natural Science Foundation of China (No. 62102205, 62076004, 62102207, 62332002, 62027804, 62088102); Anhui Provincial Key Research and Development Program under Grant 2022i01020014; The authors acknowledge the High-performance Computing Platform of Anhui University for providing computing resources.

\section*{Publish Information} 

\noindent \textbf{Name of the Journal}: Machine Intelligence Research

\noindent \textbf{DOI}:  \url{10.1007/s11633-025-1555-3}

\bibliography{reference}

%% BioMed_Central_Bib_Style_v1.01

\begin{thebibliography}{54}
% BibTex style file: bmc-mathphys.bst (version 2.1), 2014-07-24
\ifx \bisbn   \undefined \def \bisbn  #1{ISBN #1}\fi
\ifx \binits  \undefined \def \binits#1{#1}\fi
\ifx \bauthor  \undefined \def \bauthor#1{#1}\fi
\ifx \batitle  \undefined \def \batitle#1{#1}\fi
\ifx \bjtitle  \undefined \def \bjtitle#1{#1}\fi
\ifx \bvolume  \undefined \def \bvolume#1{\textbf{#1}}\fi
\ifx \byear  \undefined \def \byear#1{#1}\fi
\ifx \bissue  \undefined \def \bissue#1{#1}\fi
\ifx \bfpage  \undefined \def \bfpage#1{#1}\fi
\ifx \blpage  \undefined \def \blpage #1{#1}\fi
\ifx \burl  \undefined \def \burl#1{\textsf{#1}}\fi
\ifx \doiurl  \undefined \def \doiurl#1{\url{https://doi.org/#1}}\fi
\ifx \betal  \undefined \def \betal{\textit{et al.}}\fi
\ifx \binstitute  \undefined \def \binstitute#1{#1}\fi
\ifx \binstitutionaled  \undefined \def \binstitutionaled#1{#1}\fi
\ifx \bctitle  \undefined \def \bctitle#1{#1}\fi
\ifx \beditor  \undefined \def \beditor#1{#1}\fi
\ifx \bpublisher  \undefined \def \bpublisher#1{#1}\fi
\ifx \bbtitle  \undefined \def \bbtitle#1{#1}\fi
\ifx \bedition  \undefined \def \bedition#1{#1}\fi
\ifx \bseriesno  \undefined \def \bseriesno#1{#1}\fi
\ifx \blocation  \undefined \def \blocation#1{#1}\fi
\ifx \bsertitle  \undefined \def \bsertitle#1{#1}\fi
\ifx \bsnm \undefined \def \bsnm#1{#1}\fi
\ifx \bsuffix \undefined \def \bsuffix#1{#1}\fi
\ifx \bparticle \undefined \def \bparticle#1{#1}\fi
\ifx \barticle \undefined \def \barticle#1{#1}\fi
\bibcommenthead
\ifx \bconfdate \undefined \def \bconfdate #1{#1}\fi
\ifx \botherref \undefined \def \botherref #1{#1}\fi
\ifx \url \undefined \def \url#1{\textsf{#1}}\fi
\ifx \bchapter \undefined \def \bchapter#1{#1}\fi
\ifx \bbook \undefined \def \bbook#1{#1}\fi
\ifx \bcomment \undefined \def \bcomment#1{#1}\fi
\ifx \oauthor \undefined \def \oauthor#1{#1}\fi
\ifx \citeauthoryear \undefined \def \citeauthoryear#1{#1}\fi
\ifx \endbibitem  \undefined \def \endbibitem {}\fi
\ifx \bconflocation  \undefined \def \bconflocation#1{#1}\fi
\ifx \arxivurl  \undefined \def \arxivurl#1{\textsf{#1}}\fi
\csname PreBibitemsHook\endcsname

%%% 1
\bibitem{gallego2020eventsurvey}
\begin{barticle}
\bauthor{\bsnm{Gallego}, \binits{G.}},
\bauthor{\bsnm{Delbr{\"u}ck}, \binits{T.}},
\bauthor{\bsnm{Orchard}, \binits{G.}},
\bauthor{\bsnm{Bartolozzi}, \binits{C.}},
\bauthor{\bsnm{Taba}, \binits{B.}},
\bauthor{\bsnm{Censi}, \binits{A.}},
\bauthor{\bsnm{Leutenegger}, \binits{S.}},
\bauthor{\bsnm{Davison}, \binits{A.J.}},
\bauthor{\bsnm{Conradt}, \binits{J.}},
\bauthor{\bsnm{Daniilidis}, \binits{K.}}, \betal:
\batitle{Event-based vision: A survey}.
\bjtitle{IEEE transactions on pattern analysis and machine intelligence}
\bvolume{44}(\bissue{1}),
\bfpage{154}--\blpage{180}
(\byear{2020})
\end{barticle}
\endbibitem

%%% 2
\bibitem{wang2023eventvot}
\begin{botherref}
\oauthor{\bsnm{Wang}, \binits{X.}},
\oauthor{\bsnm{Wang}, \binits{S.}},
\oauthor{\bsnm{Tang}, \binits{C.}},
\oauthor{\bsnm{Zhu}, \binits{L.}},
\oauthor{\bsnm{Jiang}, \binits{B.}},
\oauthor{\bsnm{Tian}, \binits{Y.}},
\oauthor{\bsnm{Tang}, \binits{J.}}:
Event stream-based visual object tracking: A high-resolution benchmark dataset
  and a novel baseline.
arXiv preprint arXiv:2309.14611
(2023)
\end{botherref}
\endbibitem

%%% 3
\bibitem{wang2023visevent}
\begin{botherref}
\oauthor{\bsnm{Wang}, \binits{X.}},
\oauthor{\bsnm{Li}, \binits{J.}},
\oauthor{\bsnm{Zhu}, \binits{L.}},
\oauthor{\bsnm{Zhang}, \binits{Z.}},
\oauthor{\bsnm{Chen}, \binits{Z.}},
\oauthor{\bsnm{Li}, \binits{X.}},
\oauthor{\bsnm{Wang}, \binits{Y.}},
\oauthor{\bsnm{Tian}, \binits{Y.}},
\oauthor{\bsnm{Wu}, \binits{F.}}:
Visevent: Reliable object tracking via collaboration of frame and event flows.
IEEE Transactions on Cybernetics
(2023)
\end{botherref}
\endbibitem

%%% 4
\bibitem{li2023semantic}
\begin{botherref}
\oauthor{\bsnm{Li}, \binits{D.}},
\oauthor{\bsnm{Jin}, \binits{J.}},
\oauthor{\bsnm{Zhang}, \binits{Y.}},
\oauthor{\bsnm{Zhong}, \binits{Y.}},
\oauthor{\bsnm{Wu}, \binits{Y.}},
\oauthor{\bsnm{Chen}, \binits{L.}},
\oauthor{\bsnm{Wang}, \binits{X.}},
\oauthor{\bsnm{Luo}, \binits{B.}}:
Semantic-aware frame-event fusion based pattern recognition via large
  vision-language models.
arXiv preprint arXiv:2311.18592
(2023)
\end{botherref}
\endbibitem

%%% 5
\bibitem{wang2022hardvs}
\begin{botherref}
\oauthor{\bsnm{Wang}, \binits{X.}},
\oauthor{\bsnm{Wu}, \binits{Z.}},
\oauthor{\bsnm{Jiang}, \binits{B.}},
\oauthor{\bsnm{Bao}, \binits{Z.}},
\oauthor{\bsnm{Zhu}, \binits{L.}},
\oauthor{\bsnm{Li}, \binits{G.}},
\oauthor{\bsnm{Wang}, \binits{Y.}},
\oauthor{\bsnm{Tian}, \binits{Y.}}:
Hardvs: Revisiting human activity recognition with dynamic vision sensors.
AAAI
(2024)
\end{botherref}
\endbibitem

%%% 6
\bibitem{gehrig2023recurrentVFormer}
\begin{bchapter}
\bauthor{\bsnm{Gehrig}, \binits{M.}},
\bauthor{\bsnm{Scaramuzza}, \binits{D.}}:
\bctitle{Recurrent vision transformers for object detection with event
  cameras}.
In: \bbtitle{Proceedings of the IEEE/CVF Conference on Computer Vision and
  Pattern Recognition},
pp. \bfpage{13884}--\blpage{13893}
(\byear{2023})
\end{bchapter}
\endbibitem

%%% 7
\bibitem{teng2022nest}
\begin{bchapter}
\bauthor{\bsnm{Teng}, \binits{M.}},
\bauthor{\bsnm{Zhou}, \binits{C.}},
\bauthor{\bsnm{Lou}, \binits{H.}},
\bauthor{\bsnm{Shi}, \binits{B.}}:
\bctitle{Nest: Neural event stack for event-based image enhancement}.
In: \bbtitle{European Conference on Computer Vision},
pp. \bfpage{660}--\blpage{676}
(\byear{2022}).
\bcomment{Springer}
\end{bchapter}
\endbibitem

%%% 8
\bibitem{zhu2022eventreconstruction}
\begin{bchapter}
\bauthor{\bsnm{Zhu}, \binits{L.}},
\bauthor{\bsnm{Wang}, \binits{X.}},
\bauthor{\bsnm{Chang}, \binits{Y.}},
\bauthor{\bsnm{Li}, \binits{J.}},
\bauthor{\bsnm{Huang}, \binits{T.}},
\bauthor{\bsnm{Tian}, \binits{Y.}}:
\bctitle{Event-based video reconstruction via potential-assisted spiking neural
  network}.
In: \bbtitle{Proceedings of the IEEE/CVF Conference on Computer Vision and
  Pattern Recognition},
pp. \bfpage{3594}--\blpage{3604}
(\byear{2022})
\end{bchapter}
\endbibitem

%%% 9
\bibitem{jiang2023eventLIenhance}
\begin{botherref}
\oauthor{\bsnm{Jiang}, \binits{Y.}},
\oauthor{\bsnm{Wang}, \binits{Y.}},
\oauthor{\bsnm{Li}, \binits{S.}},
\oauthor{\bsnm{Zhang}, \binits{Y.}},
\oauthor{\bsnm{Zhao}, \binits{M.}},
\oauthor{\bsnm{Gao}, \binits{Y.}}:
Event-based low-illumination image enhancement.
IEEE Transactions on Multimedia
(2023)
\end{botherref}
\endbibitem

%%% 10
\bibitem{ding2023emlb}
\begin{botherref}
\oauthor{\bsnm{Ding}, \binits{S.}},
\oauthor{\bsnm{Chen}, \binits{J.}},
\oauthor{\bsnm{Wang}, \binits{Y.}},
\oauthor{\bsnm{Kang}, \binits{Y.}},
\oauthor{\bsnm{Song}, \binits{W.}},
\oauthor{\bsnm{Cheng}, \binits{J.}},
\oauthor{\bsnm{Cao}, \binits{Y.}}:
E-mlb: Multilevel benchmark for event-based camera denoising.
IEEE Transactions on Multimedia
(2023)
\end{botherref}
\endbibitem

%%% 11
\bibitem{wu2020probabilistic}
\begin{barticle}
\bauthor{\bsnm{Wu}, \binits{J.}},
\bauthor{\bsnm{Ma}, \binits{C.}},
\bauthor{\bsnm{Li}, \binits{L.}},
\bauthor{\bsnm{Dong}, \binits{W.}},
\bauthor{\bsnm{Shi}, \binits{G.}}:
\batitle{Probabilistic undirected graph based denoising method for dynamic
  vision sensor}.
\bjtitle{IEEE Transactions on Multimedia}
\bvolume{23},
\bfpage{1148}--\blpage{1159}
(\byear{2020})
\end{barticle}
\endbibitem

%%% 12
\bibitem{Scheerlinck2019CED:}
\begin{botherref}
\oauthor{\bsnm{Scheerlinck}, \binits{C.}},
\oauthor{\bsnm{Rebecq}, \binits{H.}},
\oauthor{\bsnm{Stoffregen}, \binits{T.}},
\oauthor{\bsnm{Barnes}, \binits{N.}},
\oauthor{\bsnm{Mahony}, \binits{R.}},
\oauthor{\bsnm{Scaramuzza}, \binits{D.}}:
Ced: Color event camera dataset.
Computing Research Repository
(2019)
\end{botherref}
\endbibitem

%%% 13
\bibitem{wang2023sstformer}
\begin{botherref}
\oauthor{\bsnm{Wang}, \binits{X.}},
\oauthor{\bsnm{Wu}, \binits{Z.}},
\oauthor{\bsnm{Rong}, \binits{Y.}},
\oauthor{\bsnm{Zhu}, \binits{L.}},
\oauthor{\bsnm{Jiang}, \binits{B.}},
\oauthor{\bsnm{Tang}, \binits{J.}},
\oauthor{\bsnm{Tian}, \binits{Y.}}:
Sstformer: Bridging spiking neural network and memory support transformer for
  frame-event based recognition.
arXiv preprint arXiv:2308.04369
(2023)
\end{botherref}
\endbibitem

%%% 14
\bibitem{jing2021VSREvent}
\begin{bchapter}
\bauthor{\bsnm{Jing}, \binits{Y.}},
\bauthor{\bsnm{Yang}, \binits{Y.}},
\bauthor{\bsnm{Wang}, \binits{X.}},
\bauthor{\bsnm{Song}, \binits{M.}},
\bauthor{\bsnm{Tao}, \binits{D.}}:
\bctitle{Turning frequency to resolution: Video super-resolution via event
  cameras}.
In: \bbtitle{Proceedings of the IEEE/CVF Conference on Computer Vision and
  Pattern Recognition},
pp. \bfpage{7772}--\blpage{7781}
(\byear{2021})
\end{bchapter}
\endbibitem

%%% 15
\bibitem{lin2019tsm}
\begin{bchapter}
\bauthor{\bsnm{Lin}, \binits{J.}},
\bauthor{\bsnm{Gan}, \binits{C.}},
\bauthor{\bsnm{Han}, \binits{S.}}:
\bctitle{Tsm: Temporal shift module for efficient video understanding}.
In: \bbtitle{Proceedings of the IEEE/CVF International Conference on Computer
  Vision},
pp. \bfpage{7083}--\blpage{7093}
(\byear{2019})
\end{bchapter}
\endbibitem

%%% 16
\bibitem{liu2021tam}
\begin{bchapter}
\bauthor{\bsnm{Liu}, \binits{Z.}},
\bauthor{\bsnm{Wang}, \binits{L.}},
\bauthor{\bsnm{Wu}, \binits{W.}},
\bauthor{\bsnm{Qian}, \binits{C.}},
\bauthor{\bsnm{Lu}, \binits{T.}}:
\bctitle{Tam: Temporal adaptive module for video recognition}.
In: \bbtitle{Proceedings of the IEEE/CVF International Conference on Computer
  Vision},
pp. \bfpage{13708}--\blpage{13718}
(\byear{2021})
\end{bchapter}
\endbibitem

%%% 17
\bibitem{liu2022video}
\begin{bchapter}
\bauthor{\bsnm{Liu}, \binits{Z.}},
\bauthor{\bsnm{Ning}, \binits{J.}},
\bauthor{\bsnm{Cao}, \binits{Y.}},
\bauthor{\bsnm{Wei}, \binits{Y.}},
\bauthor{\bsnm{Zhang}, \binits{Z.}},
\bauthor{\bsnm{Lin}, \binits{S.}},
\bauthor{\bsnm{Hu}, \binits{H.}}:
\bctitle{Video swin transformer}.
In: \bbtitle{Proceedings of the IEEE/CVF Conference on Computer Vision and
  Pattern Recognition},
pp. \bfpage{3202}--\blpage{3211}
(\byear{2022})
\end{bchapter}
\endbibitem

%%% 18
\bibitem{han2022transformersurvey}
\begin{barticle}
\bauthor{\bsnm{Han}, \binits{K.}},
\bauthor{\bsnm{Wang}, \binits{Y.}},
\bauthor{\bsnm{Chen}, \binits{H.}},
\bauthor{\bsnm{Chen}, \binits{X.}},
\bauthor{\bsnm{Guo}, \binits{J.}},
\bauthor{\bsnm{Liu}, \binits{Z.}},
\bauthor{\bsnm{Tang}, \binits{Y.}},
\bauthor{\bsnm{Xiao}, \binits{A.}},
\bauthor{\bsnm{Xu}, \binits{C.}},
\bauthor{\bsnm{Xu}, \binits{Y.}}, \betal:
\batitle{A survey on vision transformer}.
\bjtitle{IEEE transactions on pattern analysis and machine intelligence}
\bvolume{45}(\bissue{1}),
\bfpage{87}--\blpage{110}
(\byear{2022})
\end{barticle}
\endbibitem

%%% 19
\bibitem{wang2023MMPTMs}
\begin{botherref}
\oauthor{\bsnm{Wang}, \binits{X.}},
\oauthor{\bsnm{Chen}, \binits{G.}},
\oauthor{\bsnm{Qian}, \binits{G.}},
\oauthor{\bsnm{Gao}, \binits{P.}},
\oauthor{\bsnm{Wei}, \binits{X.-Y.}},
\oauthor{\bsnm{Wang}, \binits{Y.}},
\oauthor{\bsnm{Tian}, \binits{Y.}},
\oauthor{\bsnm{Gao}, \binits{W.}}:
Large-scale multi-modal pre-trained models: A comprehensive survey.
Machine Intelligence Research,
1--36
(2023)
\end{botherref}
\endbibitem

%%% 20
\bibitem{cho2023label}
\begin{bchapter}
\bauthor{\bsnm{Cho}, \binits{H.}},
\bauthor{\bsnm{Kim}, \binits{H.}},
\bauthor{\bsnm{Chae}, \binits{Y.}},
\bauthor{\bsnm{Yoon}, \binits{K.-J.}}:
\bctitle{Label-free event-based object recognition via joint learning with
  image reconstruction from events}.
In: \bbtitle{Proceedings of the IEEE/CVF International Conference on Computer
  Vision},
pp. \bfpage{19866}--\blpage{19877}
(\byear{2023})
\end{bchapter}
\endbibitem

%%% 21
\bibitem{innocenti2021temporal}
\begin{bchapter}
\bauthor{\bsnm{Innocenti}, \binits{S.U.}},
\bauthor{\bsnm{Becattini}, \binits{F.}},
\bauthor{\bsnm{Pernici}, \binits{F.}},
\bauthor{\bsnm{Del~Bimbo}, \binits{A.}}:
\bctitle{Temporal binary representation for event-based action recognition}.
In: \bbtitle{2020 25th International Conference on Pattern Recognition (ICPR)},
pp. \bfpage{10426}--\blpage{10432}
(\byear{2021}).
\bcomment{IEEE}
\end{bchapter}
\endbibitem

%%% 22
\bibitem{Kim_2022_CVPR}
\begin{bchapter}
\bauthor{\bsnm{Kim}, \binits{J.}},
\bauthor{\bsnm{Hwang}, \binits{I.}},
\bauthor{\bsnm{Kim}, \binits{Y.M.}}:
\bctitle{Ev-tta: Test-time adaptation for event-based object recognition}.
In: \bbtitle{Proceedings of the IEEE/CVF Conference on Computer Vision and
  Pattern Recognition (CVPR)},
pp. \bfpage{17745}--\blpage{17754}
(\byear{2022})
\end{bchapter}
\endbibitem

%%% 23
\bibitem{cannici2019attention}
\begin{bchapter}
\bauthor{\bsnm{Cannici}, \binits{M.}},
\bauthor{\bsnm{Ciccone}, \binits{M.}},
\bauthor{\bsnm{Romanoni}, \binits{A.}},
\bauthor{\bsnm{Matteucci}, \binits{M.}}:
\bctitle{Attention mechanisms for object recognition with event-based cameras}.
In: \bbtitle{2019 IEEE Winter Conference on Applications of Computer Vision
  (WACV)},
pp. \bfpage{1127}--\blpage{1136}
(\byear{2019}).
\bcomment{IEEE}
\end{bchapter}
\endbibitem

%%% 24
\bibitem{Wang_2019_CVPR}
\begin{bchapter}
\bauthor{\bsnm{Wang}, \binits{Y.}},
\bauthor{\bsnm{Du}, \binits{B.}},
\bauthor{\bsnm{Shen}, \binits{Y.}},
\bauthor{\bsnm{Wu}, \binits{K.}},
\bauthor{\bsnm{Zhao}, \binits{G.}},
\bauthor{\bsnm{Sun}, \binits{J.}},
\bauthor{\bsnm{Wen}, \binits{H.}}:
\bctitle{Ev-gait: Event-based robust gait recognition using dynamic vision
  sensors}.
In: \bbtitle{Proceedings of the IEEE/CVF Conference on Computer Vision and
  Pattern Recognition (CVPR)}
(\byear{2019})
\end{bchapter}
\endbibitem

%%% 25
\bibitem{liu2021event}
\begin{bchapter}
\bauthor{\bsnm{Liu}, \binits{Q.}},
\bauthor{\bsnm{Xing}, \binits{D.}},
\bauthor{\bsnm{Tang}, \binits{H.}},
\bauthor{\bsnm{Ma}, \binits{D.}},
\bauthor{\bsnm{Pan}, \binits{G.}}:
\bctitle{Event-based action recognition using motion information and spiking
  neural networks.}
In: \bbtitle{IJCAI},
pp. \bfpage{1743}--\blpage{1749}
(\byear{2021})
\end{bchapter}
\endbibitem

%%% 26
\bibitem{islam2022maven}
\begin{botherref}
\oauthor{\bsnm{Islam}, \binits{M.M.}},
\oauthor{\bsnm{Yasar}, \binits{M.S.}},
\oauthor{\bsnm{Iqbal}, \binits{T.}}:
Maven: A memory augmented recurrent approach for multimodal fusion.
IEEE Transactions on Multimedia
(2022)
\end{botherref}
\endbibitem

%%% 27
\bibitem{mai2022multimodalbottleneck}
\begin{botherref}
\oauthor{\bsnm{Mai}, \binits{S.}},
\oauthor{\bsnm{Zeng}, \binits{Y.}},
\oauthor{\bsnm{Hu}, \binits{H.}}:
Multimodal information bottleneck: Learning minimal sufficient unimodal and
  multimodal representations.
IEEE Transactions on Multimedia
(2022)
\end{botherref}
\endbibitem

%%% 28
\bibitem{wang2022mfgnet}
\begin{botherref}
\oauthor{\bsnm{Wang}, \binits{X.}},
\oauthor{\bsnm{Shu}, \binits{X.}},
\oauthor{\bsnm{Zhang}, \binits{S.}},
\oauthor{\bsnm{Jiang}, \binits{B.}},
\oauthor{\bsnm{Wang}, \binits{Y.}},
\oauthor{\bsnm{Tian}, \binits{Y.}},
\oauthor{\bsnm{Wu}, \binits{F.}}:
Mfgnet: Dynamic modality-aware filter generation for rgb-t tracking.
IEEE Transactions on Multimedia
(2022)
\end{botherref}
\endbibitem

%%% 29
\bibitem{huang2022vefnet}
\begin{bchapter}
\bauthor{\bsnm{Huang}, \binits{Z.}},
\bauthor{\bsnm{Huang}, \binits{R.}},
\bauthor{\bsnm{Sun}, \binits{L.}},
\bauthor{\bsnm{Zhao}, \binits{C.}},
\bauthor{\bsnm{Huang}, \binits{M.}},
\bauthor{\bsnm{Su}, \binits{S.}}:
\bctitle{Vefnet: An event-rgb cross modality fusion network for visual place
  recognition}.
In: \bbtitle{2022 IEEE International Conference on Image Processing (ICIP)},
pp. \bfpage{2671}--\blpage{2675}
(\byear{2022}).
\bcomment{IEEE}
\end{bchapter}
\endbibitem

%%% 30
\bibitem{vaswani2017attention}
\begin{botherref}
\oauthor{\bsnm{Vaswani}, \binits{A.}},
\oauthor{\bsnm{Shazeer}, \binits{N.}},
\oauthor{\bsnm{Parmar}, \binits{N.}},
\oauthor{\bsnm{Uszkoreit}, \binits{J.}},
\oauthor{\bsnm{Jones}, \binits{L.}},
\oauthor{\bsnm{Gomez}, \binits{A.N.}},
\oauthor{\bsnm{Kaiser}, \binits{{\L}.}},
\oauthor{\bsnm{Polosukhin}, \binits{I.}}:
Attention is all you need.
Advances in neural information processing systems
\textbf{30}
(2017)
\end{botherref}
\endbibitem

%%% 31
\bibitem{Truong_2021_ICCV}
\begin{bchapter}
\bauthor{\bsnm{Truong}, \binits{T.-D.}},
\bauthor{\bsnm{Duong}, \binits{C.N.}},
\bauthor{\bsnm{De~Vu}, \binits{T.}},
\bauthor{\bsnm{Pham}, \binits{H.A.}},
\bauthor{\bsnm{Raj}, \binits{B.}},
\bauthor{\bsnm{Le}, \binits{N.}},
\bauthor{\bsnm{Luu}, \binits{K.}}:
\bctitle{The right to talk: An audio-visual transformer approach}.
In: \bbtitle{Proceedings of the IEEE/CVF International Conference on Computer
  Vision (ICCV)},
pp. \bfpage{1105}--\blpage{1114}
(\byear{2021})
\end{bchapter}
\endbibitem

%%% 32
\bibitem{radford2021learning}
\begin{bchapter}
\bauthor{\bsnm{Radford}, \binits{A.}},
\bauthor{\bsnm{Kim}, \binits{J.W.}},
\bauthor{\bsnm{Hallacy}, \binits{C.}},
\bauthor{\bsnm{Ramesh}, \binits{A.}},
\bauthor{\bsnm{Goh}, \binits{G.}},
\bauthor{\bsnm{Agarwal}, \binits{S.}},
\bauthor{\bsnm{Sastry}, \binits{G.}},
\bauthor{\bsnm{Askell}, \binits{A.}},
\bauthor{\bsnm{Mishkin}, \binits{P.}},
\bauthor{\bsnm{Clark}, \binits{J.}}, \betal:
\bctitle{Learning transferable visual models from natural language
  supervision}.
In: \bbtitle{International Conference on Machine Learning},
pp. \bfpage{8748}--\blpage{8763}
(\byear{2021}).
\bcomment{PMLR}
\end{bchapter}
\endbibitem

%%% 33
\bibitem{He_2022_CVPR}
\begin{bchapter}
\bauthor{\bsnm{He}, \binits{K.}},
\bauthor{\bsnm{Chen}, \binits{X.}},
\bauthor{\bsnm{Xie}, \binits{S.}},
\bauthor{\bsnm{Li}, \binits{Y.}},
\bauthor{\bsnm{Doll\'ar}, \binits{P.}},
\bauthor{\bsnm{Girshick}, \binits{R.}}:
\bctitle{Masked autoencoders are scalable vision learners}.
In: \bbtitle{Proceedings of the IEEE/CVF Conference on Computer Vision and
  Pattern Recognition (CVPR)},
pp. \bfpage{16000}--\blpage{16009}
(\byear{2022})
\end{bchapter}
\endbibitem

%%% 34
\bibitem{Neimark_2021_ICCV}
\begin{bchapter}
\bauthor{\bsnm{Neimark}, \binits{D.}},
\bauthor{\bsnm{Bar}, \binits{O.}},
\bauthor{\bsnm{Zohar}, \binits{M.}},
\bauthor{\bsnm{Asselmann}, \binits{D.}}:
\bctitle{Video transformer network}.
In: \bbtitle{Proceedings of the IEEE/CVF International Conference on Computer
  Vision (ICCV) Workshops},
pp. \bfpage{3163}--\blpage{3172}
(\byear{2021})
\end{bchapter}
\endbibitem

%%% 35
\bibitem{peng2021conformer}
\begin{bchapter}
\bauthor{\bsnm{Peng}, \binits{Z.}},
\bauthor{\bsnm{Huang}, \binits{W.}},
\bauthor{\bsnm{Gu}, \binits{S.}},
\bauthor{\bsnm{Xie}, \binits{L.}},
\bauthor{\bsnm{Wang}, \binits{Y.}},
\bauthor{\bsnm{Jiao}, \binits{J.}},
\bauthor{\bsnm{Ye}, \binits{Q.}}:
\bctitle{Conformer: Local features coupling global representations for visual
  recognition}.
In: \bbtitle{Proceedings of the IEEE/CVF International Conference on Computer
  Vision},
pp. \bfpage{367}--\blpage{376}
(\byear{2021})
\end{bchapter}
\endbibitem

%%% 36
\bibitem{nagrani2021AttBottleNeck}
\begin{barticle}
\bauthor{\bsnm{Nagrani}, \binits{A.}},
\bauthor{\bsnm{Yang}, \binits{S.}},
\bauthor{\bsnm{Arnab}, \binits{A.}},
\bauthor{\bsnm{Jansen}, \binits{A.}},
\bauthor{\bsnm{Schmid}, \binits{C.}},
\bauthor{\bsnm{Sun}, \binits{C.}}:
\batitle{Attention bottlenecks for multimodal fusion}.
\bjtitle{Advances in Neural Information Processing Systems}
\bvolume{34},
\bfpage{14200}--\blpage{14213}
(\byear{2021})
\end{barticle}
\endbibitem

%%% 37
\bibitem{zhang2023cmx}
\begin{botherref}
\oauthor{\bsnm{Zhang}, \binits{J.}},
\oauthor{\bsnm{Liu}, \binits{H.}},
\oauthor{\bsnm{Yang}, \binits{K.}},
\oauthor{\bsnm{Hu}, \binits{X.}},
\oauthor{\bsnm{Liu}, \binits{R.}},
\oauthor{\bsnm{Stiefelhagen}, \binits{R.}}:
Cmx: Cross-modal fusion for rgb-x semantic segmentation with transformers.
IEEE Transactions on Intelligent Transportation Systems
(2023)
\end{botherref}
\endbibitem

%%% 38
\bibitem{zhao2022tftn}
\begin{barticle}
\bauthor{\bsnm{Zhao}, \binits{C.}},
\bauthor{\bsnm{Liu}, \binits{H.}},
\bauthor{\bsnm{Su}, \binits{N.}},
\bauthor{\bsnm{Yan}, \binits{Y.}}:
\batitle{Tftn: A transformer-based fusion tracking framework of hyperspectral
  and rgb}.
\bjtitle{IEEE Transactions on Geoscience and Remote Sensing}
\bvolume{60},
\bfpage{1}--\blpage{15}
(\byear{2022})
\end{barticle}
\endbibitem

%%% 39
\bibitem{bozic2021transformerfusion}
\begin{barticle}
\bauthor{\bsnm{Bozic}, \binits{A.}},
\bauthor{\bsnm{Palafox}, \binits{P.}},
\bauthor{\bsnm{Thies}, \binits{J.}},
\bauthor{\bsnm{Dai}, \binits{A.}},
\bauthor{\bsnm{Nie{\ss}ner}, \binits{M.}}:
\batitle{Transformerfusion: Monocular rgb scene reconstruction using
  transformers}.
\bjtitle{Advances in Neural Information Processing Systems}
\bvolume{34},
\bfpage{1403}--\blpage{1414}
(\byear{2021})
\end{barticle}
\endbibitem

%%% 40
\bibitem{wang2021transformer}
\begin{botherref}
\oauthor{\bsnm{Wang}, \binits{Y.}},
\oauthor{\bsnm{Jia}, \binits{X.}},
\oauthor{\bsnm{Zhang}, \binits{L.}},
\oauthor{\bsnm{Li}, \binits{Y.}},
\oauthor{\bsnm{Elder}, \binits{J.}},
\oauthor{\bsnm{Lu}, \binits{H.}}:
Transformer-based network for rgb-d saliency detection.
arXiv preprint arXiv:2112.00582
(2021)
\end{botherref}
\endbibitem

%%% 41
\bibitem{zhou2022multispectral}
\begin{barticle}
\bauthor{\bsnm{Zhou}, \binits{H.}},
\bauthor{\bsnm{Tian}, \binits{C.}},
\bauthor{\bsnm{Zhang}, \binits{Z.}},
\bauthor{\bsnm{Huo}, \binits{Q.}},
\bauthor{\bsnm{Xie}, \binits{Y.}},
\bauthor{\bsnm{Li}, \binits{Z.}}:
\batitle{Multispectral fusion transformer network for rgb-thermal urban scene
  semantic segmentation}.
\bjtitle{IEEE Geoscience and Remote Sensing Letters}
\bvolume{19},
\bfpage{1}--\blpage{5}
(\byear{2022})
\end{barticle}
\endbibitem

%%% 42
\bibitem{li2021trear}
\begin{barticle}
\bauthor{\bsnm{Li}, \binits{X.}},
\bauthor{\bsnm{Hou}, \binits{Y.}},
\bauthor{\bsnm{Wang}, \binits{P.}},
\bauthor{\bsnm{Gao}, \binits{Z.}},
\bauthor{\bsnm{Xu}, \binits{M.}},
\bauthor{\bsnm{Li}, \binits{W.}}:
\batitle{Trear: Transformer-based rgb-d egocentric action recognition}.
\bjtitle{IEEE Transactions on Cognitive and Developmental Systems}
\bvolume{14}(\bissue{1}),
\bfpage{246}--\blpage{252}
(\byear{2021})
\end{barticle}
\endbibitem

%%% 43
\bibitem{Wang_2022_CVPR}
\begin{bchapter}
\bauthor{\bsnm{Wang}, \binits{Y.}},
\bauthor{\bsnm{Chen}, \binits{X.}},
\bauthor{\bsnm{Cao}, \binits{L.}},
\bauthor{\bsnm{Huang}, \binits{W.}},
\bauthor{\bsnm{Sun}, \binits{F.}},
\bauthor{\bsnm{Wang}, \binits{Y.}}:
\bctitle{Multimodal token fusion for vision transformers}.
In: \bbtitle{Proceedings of the IEEE/CVF Conference on Computer Vision and
  Pattern Recognition (CVPR)},
pp. \bfpage{12186}--\blpage{12195}
(\byear{2022})
\end{bchapter}
\endbibitem

%%% 44
\bibitem{Prakash_2021_CVPR}
\begin{bchapter}
\bauthor{\bsnm{Prakash}, \binits{A.}},
\bauthor{\bsnm{Chitta}, \binits{K.}},
\bauthor{\bsnm{Geiger}, \binits{A.}}:
\bctitle{Multi-modal fusion transformer for end-to-end autonomous driving}.
In: \bbtitle{Proceedings of the IEEE/CVF Conference on Computer Vision and
  Pattern Recognition (CVPR)},
pp. \bfpage{7077}--\blpage{7087}
(\byear{2021})
\end{bchapter}
\endbibitem

%%% 45
\bibitem{He_2016_CVPR}
\begin{bchapter}
\bauthor{\bsnm{He}, \binits{K.}},
\bauthor{\bsnm{Zhang}, \binits{X.}},
\bauthor{\bsnm{Ren}, \binits{S.}},
\bauthor{\bsnm{Sun}, \binits{J.}}:
\bctitle{Deep residual learning for image recognition}.
In: \bbtitle{Proceedings of the IEEE Conference on Computer Vision and Pattern
  Recognition (CVPR)}
(\byear{2016})
\end{bchapter}
\endbibitem

%%% 46
\bibitem{dosovitskiy2020image}
\begin{bchapter}
\bauthor{\bsnm{Dosovitskiy}, \binits{A.}},
\bauthor{\bsnm{Beyer}, \binits{L.}},
\bauthor{\bsnm{Kolesnikov}, \binits{A.}},
\bauthor{\bsnm{Weissenborn}, \binits{D.}},
\bauthor{\bsnm{Zhai}, \binits{X.}},
\bauthor{\bsnm{Unterthiner}, \binits{T.}},
\bauthor{\bsnm{Dehghani}, \binits{M.}},
\bauthor{\bsnm{Minderer}, \binits{M.}},
\bauthor{\bsnm{Heigold}, \binits{G.}},
\bauthor{\bsnm{Gelly}, \binits{S.}}, \betal:
\bctitle{An image is worth 16x16 words: Transformers for image recognition at
  scale}.
In: \bbtitle{International Conference on Learning Representations}
(\byear{2020})
\end{bchapter}
\endbibitem

%%% 47
\bibitem{Bottou2012Stochastic}
\begin{botherref}
\oauthor{\bsnm{Bottou}, \binits{L.}}:
Stochastic gradient descent tricks.
Neural Networks: Tricks of the Trade (2nd ed.)
(2012)
\end{botherref}
\endbibitem

%%% 48
\bibitem{paszke2019pytorch}
\begin{botherref}
\oauthor{\bsnm{Paszke}, \binits{A.}},
\oauthor{\bsnm{Gross}, \binits{S.}},
\oauthor{\bsnm{Massa}, \binits{F.}},
\oauthor{\bsnm{Lerer}, \binits{A.}},
\oauthor{\bsnm{Bradbury}, \binits{J.}},
\oauthor{\bsnm{Chanan}, \binits{G.}},
\oauthor{\bsnm{Killeen}, \binits{T.}},
\oauthor{\bsnm{Lin}, \binits{Z.}},
\oauthor{\bsnm{Gimelshein}, \binits{N.}},
\oauthor{\bsnm{Antiga}, \binits{L.}}, et al.:
Pytorch: An imperative style, high-performance deep learning library.
Advances in neural information processing systems
\textbf{32}
(2019)
\end{botherref}
\endbibitem

%%% 49
\bibitem{tran2015learning}
\begin{bchapter}
\bauthor{\bsnm{Tran}, \binits{D.}},
\bauthor{\bsnm{Bourdev}, \binits{L.}},
\bauthor{\bsnm{Fergus}, \binits{R.}},
\bauthor{\bsnm{Torresani}, \binits{L.}},
\bauthor{\bsnm{Paluri}, \binits{M.}}:
\bctitle{Learning spatiotemporal features with 3d convolutional networks}.
In: \bbtitle{Proceedings of the IEEE International Conference on Computer
  Vision},
pp. \bfpage{4489}--\blpage{4497}
(\byear{2015})
\end{bchapter}
\endbibitem

%%% 50
\bibitem{feichtenhofer2020x3d}
\begin{bchapter}
\bauthor{\bsnm{Feichtenhofer}, \binits{C.}}:
\bctitle{X3d: Expanding architectures for efficient video recognition}.
In: \bbtitle{Proceedings of the IEEE/CVF Conference on Computer Vision and
  Pattern Recognition},
pp. \bfpage{203}--\blpage{213}
(\byear{2020})
\end{bchapter}
\endbibitem

%%% 51
\bibitem{bertasius2021space}
\begin{bchapter}
\bauthor{\bsnm{Bertasius}, \binits{G.}},
\bauthor{\bsnm{Wang}, \binits{H.}},
\bauthor{\bsnm{Torresani}, \binits{L.}}:
\bctitle{Is space-time attention all you need for video understanding?}
In: \bbtitle{ICML},
vol. \bseriesno{2},
p. \bfpage{4}
(\byear{2021})
\end{bchapter}
\endbibitem

%%% 52
\bibitem{li2022mvitv2}
\begin{bchapter}
\bauthor{\bsnm{Li}, \binits{Y.}},
\bauthor{\bsnm{Wu}, \binits{C.-Y.}},
\bauthor{\bsnm{Fan}, \binits{H.}},
\bauthor{\bsnm{Mangalam}, \binits{K.}},
\bauthor{\bsnm{Xiong}, \binits{B.}},
\bauthor{\bsnm{Malik}, \binits{J.}},
\bauthor{\bsnm{Feichtenhofer}, \binits{C.}}:
\bctitle{Mvitv2: Improved multiscale vision transformers for classification and
  detection}.
In: \bbtitle{Proceedings of the IEEE/CVF Conference on Computer Vision and
  Pattern Recognition},
pp. \bfpage{4804}--\blpage{4814}
(\byear{2022})
\end{bchapter}
\endbibitem

%%% 53
\bibitem{wang2021action}
\begin{bchapter}
\bauthor{\bsnm{Wang}, \binits{Z.}},
\bauthor{\bsnm{She}, \binits{Q.}},
\bauthor{\bsnm{Smolic}, \binits{A.}}:
\bctitle{Action-net: Multipath excitation for action recognition}.
In: \bbtitle{Proceedings of the IEEE/CVF Conference on Computer Vision and
  Pattern Recognition},
pp. \bfpage{13214}--\blpage{13223}
(\byear{2021})
\end{bchapter}
\endbibitem

%%% 54
\bibitem{li2023blip2}
\begin{botherref}
\oauthor{\bsnm{Li}, \binits{J.}},
\oauthor{\bsnm{Li}, \binits{D.}},
\oauthor{\bsnm{Savarese}, \binits{S.}},
\oauthor{\bsnm{Hoi}, \binits{S.}}:
Blip-2: Bootstrapping language-image pre-training with frozen image encoders
  and large language models
(2023)
\end{botherref}
\endbibitem

\end{thebibliography}

\end{document}